%% file: main.tex
\documentclass[letterpaper]{article} 
\usepackage{aaai24}  
\usepackage{times}  
\usepackage{helvet}  
\usepackage{courier}  
\usepackage[hyphens]{url}  
\usepackage{graphicx} 
\usepackage{natbib}  
\usepackage{caption} 
\usepackage[algo2e]{algorithm2e}
\usepackage{algorithm}
\usepackage{algorithmic}
\usepackage{newfloat}
\usepackage{listings}
\DeclareCaptionStyle{ruled}{labelfont=normalfont,labelsep=colon,strut=off} 
\floatstyle{ruled}
\newfloat{listing}{tb}{lst}{}
\floatname{listing}{Listing}

\usepackage{subfigure}
\PassOptionsToPackage{numbers,sort&compress}{natbib}
\usepackage{nicefrac}
\usepackage{amssymb}
\usepackage{amsthm}
\usepackage{newtxmath}
\usepackage{scrextend}
\usepackage{numprint}
\usepackage{makecell}
\usepackage{pbox}
\usepackage{xurl}

\newtheorem{theorem}{Theorem}[section]

\newtheorem{lemma}[theorem]{Lemma}
\newtheorem*{lemma*}{Lemma}

\npthousandsep{,}

\usepackage[utf8]{inputenc} 
\usepackage[T1]{fontenc}    
\usepackage{url}            
\usepackage{booktabs}       
\usepackage{amsfonts}       
\usepackage{nicefrac}       
\usepackage{microtype}      
\usepackage{xcolor}         
\usepackage{xspace}
\usepackage{xurl}

\newcommand{\reals}{\mathbb{R}}
\newcommand{\f}{\mathbf{f}}
\newcommand{\x}{\mathbf{x}}

\newcommand{\att}{S} 

\newcommand{\pointer}{*}

\newcommand{\Secref}[1]{Section~\ref{#1}}
\newcommand{\Eqref}[1]{Eq.~\ref{#1}}
\newcommand{\Figref}[1]{Figure~\ref{#1}}

\newcommand{\Lemref}[1]{Lemma~\ref{#1}}
\newcommand{\agg}{\mbox{AGG}}

\newcommand{\acronym}{TREE-G\xspace}

\graphicspath{ {images/} }

\title{TREE-G: Decision Trees Contesting Graph Neural Networks}
\author {
    Maya Bechler-Speicher\textsuperscript{\rm 1},
    Amir Globerson\textsuperscript{\rm 1},
    Ran Gilad-Bachrach\textsuperscript{\rm 2}
}
\affiliations {
    \textsuperscript{\rm 1} Blavatnik School of Computer Science, Tel-Aviv University\\
    \textsuperscript{\rm 2} Department of Bio-Medical Engineering and Edmond J. Safra Center for Bioinformatics,Tel-Aviv University\\
    mayab4@mail.tau.ac.il
     }

\begin{document}
\maketitle
\begin{abstract}
When dealing with tabular data, models based on decision trees are a popular choice due to their high accuracy on these data types, their ease of application, and explainability properties. 
However, when it comes to graph-structured data, it is not clear how to apply them effectively, in a way that incorporates the topological information with the tabular data available on the vertices of the graph.
To address this challenge, we introduce \acronym. 
\acronym modifies standard decision trees, by introducing a novel split function that is specialized for graph data. Not only does this split function incorporate the node features and the topological information, but it also uses a novel pointer mechanism that allows split nodes to use information computed in previous splits. Therefore, the split function adapts to the predictive task and the graph at hand. 
We analyze the theoretical properties of \acronym 
and demonstrate its benefits empirically on multiple graph and vertex prediction benchmarks. In these experiments, \acronym consistently outperforms other tree-based models and often outperforms  other graph-learning algorithms such as Graph Neural Networks and Graph Kernels, sometimes by large margins. Moreover, \acronym{}s models and their predictions can be explained and visualized.

\end{abstract}

\section{Introduction}
\input{intro.tex}

\section{Related Work}{\label{Section:relatedwork}}
\input{related_work.tex}

\section{TREE-G}{\label{Section:methods}}
\input{method.tex}

\section{Theoretical Properties of \acronym}\label{sec:theoretical_analysis}
\input{theoretical_properties.tex}

\section{Empirical Evaluation}{\label{Section:experiments}}
\input{experiments.tex}

\section{Explaining \acronym Outputs}{\label{Section:explainability}}
\input{explanations.tex}

\section{Conclusion}\label{sec:conclusions}
\input{conclusion.tex}

\section*{Acknowledgments}
This work was supported by a grant from the Tel Aviv University Center for AI and Data Science (TAD) and by the Israeli Science Foundation research grant 1186/18 and 1437/22.

\bibliography{references}

\clearpage
\appendix
\onecolumn

\input{appendix}



\end{document}

%% file: intro.tex
Decision Trees (DTs) are one of the cornerstones of modern data science and are a leading alternative to neural networks, especially in tabular data settings, where DTs often outperform deep learning models~\cite{shwartz-ziv2021tabular, nnvsgbt, inbook, treesarebetterfortabulardata}. DTs are popular with practitioners due to their ease of use, out-of-the-box high performance, and explainability properties. In a large survey conducted by Kaggle\footnote{\url{www.kaggle.com/competitions/kaggle-survey-2022/data}} in 2022  with  $23{,}997$ data scientists worldwide, $63\%$ of the participants reported using decision trees, Random Forests (RF) or Gradient Boosted Trees (GBT) on a regular basis~\cite{rf, friedman2000greedy, StochasticGradientBoosting}.

In this study we investigate whether the merits of DTs can be repeated in tasks that involve \emph{graph} structured data. Such tasks are found in diverse domains including social studies, molecular biology, drug discovery, and communication~\cite{sperduti1997supervised, gori2005}.
The ability to perform regression and classification on graphs is important in these domains and others, which explains the ongoing efforts to develop machine learning algorithms for graphs~\cite{gratedgraph, inductivegraph, mulecularfingerprints, cnngnn, barrett2018measuring, geomodel, rossi2020temporal, generative}.

Graph data often exhibits characteristics similar to those of tabular data. For instance, in a social network, vertices represent individuals with associated features such as age, marital status, and education. Given the efficacy of DTs with tabular data, one might wonder if this success can be translated to graph-related tasks. However, adapting DTs to integrate both the graph structure and vertex attributes remains a challenge. To address this gap, we present a novel and efficient technique, named \acronym, to apply DTs to graphs.


DTs split the input space by comparing feature values to thresholds. When operating on graphs, these split decisions should incorporate the features on the vertices of the graph as well as its topological structure.\footnote{For the sake of clarity, we ignore edge-features in this work.} Moreover, these decisions should be applicable to graphs of varying sizes and satisfy the appropriate invariance and equivariance properties with respect to vertex ordering. Current methods achieve these goals using 'feature engineering', a process in which the ``raw features'' are augmented with computed features using pre-defined functions that may use the graph structure. These features, which are computed before the evaluation (or training) of the DT, use domain knowledge, or graph theoretic measures~\cite{barnett2016feature, de2018simple}. However, these approaches are labor intensive, require domain knowledge, and may fail to capture interactions between vertex features and graph structure. We show this formally in \Lemref{thm:GTA_ge_dt} where a pair of non-isomorphic graphs over $4$ vertices and $2$ features are shown to be inseparable by standard tree methods.

\acronym enhances standard decision trees by introducing a novel split function, specialized for graph data which is dynamically evaluated while the DT is applied to a graph. It supports both directed and undirected graphs of arbitrary sizes and can be used for classification, regression, and other prediction tasks on entire graphs or on the vertices and edges of these graphs. It preserves the standard decision trees framework and thus can be used in any ensemble method, for instance, as a weak learner in GBT. Indeed, we show in an extensive empirical evaluation that \acronym with GBT generates models that are superior to other tree-based methods. Furthermore, it often outperforms Graph Neural Networks (GNNs), the leading deep-learning method for learning over graphs. 
In some cases, the difference in performance between \acronym and the other methods reaches $\sim 6.4$ percentage points.

A key idea in \acronym is that split functions in the tree should be able to focus on a subset of the nodes in the graph to identify relevant substructures. For example, in the task of gauging the tendency of a molecule to increase genetic mutation frequency this should allow models to identify substructures such as carbon rings (cycles), that are known to be mutagenic~\cite{mutagenicaffect}. When task-related substructures are known beforehand, one can compute task-specific features based on them. However, when there's a need to learn these substructures, finding a method that is computationally efficient and consistent across different graphs remains a challenge.
\acronym addresses this challenge by a novel mechanism to dynamically generate candidate subsets of vertices at each split-node as the graph traverses the tree. These subsets can be used in split functions of downstream split-nodes combined with the vertex features and the topology of the graph. 
Ablation studies we performed confirmed that this ability is a key ingredient of the empirical superiority of \acronym over other models.
As we prove later, \acronym is more expressive than standard decision trees that only act on the input features, even if some topological features which are computed in advance are added as input features.

\textbf{The main contributions} of this work are: 
(1) We present \acronym, a novel decision tree model specialized for graph data.
(2) We empirically show that \acronym based ensembles outperform existing tree-based methods and graph kernels. Furthermore, \acronym is very competitive with GNNs and often outperforms them.
(3) We provide theoretical results that highlight the properties and expressive power of \acronym.
In addition, we introduce an explainability mechanism for \acronym.

%% file: related_work.tex

\textbf{Machine Learning on Graphs} Many methods for learning on graphs were proposed. In Graph Kernels, a predefined embedding is used to represent graphs in a vector space to which cosine similarities are applied~\cite{rw,wl,gk}. A more recent line of work utilizes deep-learning approaches:
GNNs learn a representation vector for each vertex of a given graph using an iterative neighborhood aggregation process. These representations are used in downstream tasks~\cite{gori2005, mpgnn, pinsage}. 
See \citet{wu2020comprehensive, gksurvey} for recent surveys.

\textbf{DTs for Graph Learning} 
Several approaches have been introduced to allow DTs to operate on graphs~\cite{barnett2016feature, de2018simple,heaton2016empirical, he2017simboost, lei2019gbdtcda, 8530225, article}.
These approaches combine domain-specific feature engineering with graph-theoretic features.
Other approaches combine GNNs with DTs in various ways and have mostly been designed for specific problems~\cite{10.1145/3292500.3330676}. For example, XGraphBoost was used for a drug discovery task~\cite{xgraphboost, STOKES2020688}. It uses GNNs to embed the graph into a vector and applies GBT~\cite{friedman2000greedy, StochasticGradientBoosting} to make predictions using this vector representation. A similar method has been used for link prediction between human phenotypes and genes~\cite{patel2021graph}. The recently proposed BGNN model predicts vertex attributes by alternating between training GBT and GNNs~\cite{ivanov2021boost}. Thus, existing solutions either combine deep-learning components or focus on specific domains utilizing domain expertise.

\acronym, detailed in the subsequent section, offers a 'pure' and generic DT solution tailored for graph-based tasks. It eliminates the need for domain knowledge or neural networks, yet remains competitive with these approaches.
It is inspired by Set-Trees~\cite{settrees} where DTs were extended to operate on data that had a set structure by introducing an attention mechanism. Sets can be viewed as graphs with no edges. However, note that since Set-Trees are designed to work on sets, they do not accommodate the topological structure of graphs nor handle the diverse types of tasks that graph data gives rise to such as vertex or link labeling

%% file: method.tex
In this section we introduce the \acronym model.



\subsection{Preliminaries}

A graph $G$ is defined by a set of vertices $V$ of size $n$ and an adjacency matrix $A$ of size $n\times n$ that can be symmetric or a-symmetric. The entry $A_{ij}$ indicates an edge from vertex $j$ to vertex $i$.
Each vertex is associated with a vector of $l$ real-valued features, and we denote the stacked matrix of feature vectors over all the vertices by $X$. We denote the $k$'th column of this matrix by $\f_k$; i.e., entry $i$ in  $\f_k$ holds the value of the $k$\textsuperscript{th} feature of vertex $i$. The $i$\textsuperscript{th} entry of a vector $\x$ is denoted by $x_i$ or $(\x)_i$. 
Recall that $\left(A^d\right)_{ij}$ is the number of walks of length $d$ between vertex $j$ to vertex $i$.
We focus on two tasks: \textit{graph labeling} and \textit{vertex labeling} (each includes both classification and regression). In the former, the goal is to assign a label to the entire graph while in the latter, the goal is to assign a label to individual vertices.\footnote{Predicting properties of edges, can be performed using the dual line graph as described in the Appendix.}
In vertex labeling tasks, \acronym takes in a graph along with one of its vertices and predicts the label of that vertex. In contrast, for graph labeling tasks, \acronym is fed with a graph and predicts the label for the entire graph.
Given the variety of graph tasks, \acronym comes in multiple variants.
It's important to note that trees are also considered graphs. To prevent any confusion, we use terms like \textit{root, node, split-node, leaf} to refer to decision tree nodes, and \textit{vertex, vertices} for graph vertices.




\subsection{A Split Function Specialized for Graphs}
For the sake of clarity, we begin by highlighting the parts of the standard DTs framework TREE-G modifies. 
During the inference phase in standard DTs, a vector example $\x$ traverses the tree until a leaf is reached and the value stored in the leaf is reported. At each split-node, the $k$'th feature of $\x$ is compared against a threshold, $x_k>\theta$. During training, the goal is to select in each split-node, the optimal feature index $k$ and threshold value $\theta$, with respect to some optimization criterion, e.g., Gini score or the $L_2$-loss.
Training is conducted greedily by converting a leaf into a decision node with two new leaves to optimize the criterion.

\begin{algorithm}[t]
\KwData{$A$: adjacency matrix, $X$: node features matrix, $i$: node index for inference in the case of vertex labeling task, $N$: a TREE-G node with stored parameters $k_N,d_N,\pointer_N,\rho_N, r_N, \theta_N$and $\agg_N$ in the case of graph labeling tasks}
\KwResult{label  $y$}
\SetKwFunction{InferVertex}{InferVertex}
\SetKwFunction{InferGraph}{InferGraph}
\SetKwProg{Fn}{Function}{:}{}

\Fn{\InferVertex{$N$, $A$, $X$, $i$}}{
    \If{$N$ is a leaf node}{
        \Return the label stored in $N$\;
    }

    Evaluate $\phi$  = $\phi_{k_N,d_N,\pointer_N,\rho_N, r_N}(A, X, i)$ \;
    
    \If{$\phi > \theta_N$}{
        \Return \InferVertex{$N_{ge}$, $A$, $X$, $i$} where $N_{ge}$ is a child of $N$ \;
    }
    \Else{
    \Return \InferVertex{$N_{leq}$, $A$, $X$, $i$} where $N_{leq}$ is a child of $N$ \;
    }

}
\Fn{\InferGraph{$N$, $G$}}{
    \If{$N$ is a leaf node}{
        \Return the label stored in $N$\;
    }

   Evaluate $\phi$  = $\phi_{k_N,d_N,\pointer_N,\rho_N, r_N, \agg_N}(A, X)$ \;
    
    \If{$\phi > \theta_N$}{
        \Return \InferGraph{$N_{ge}$, $G$} where $N_{ge}$ is a child of $N$ \;
    }
    \Else{
    \Return  \InferGraph{$N_{leq}$, $A$, $X$} where $N_{leq}$ is a child of $N$ \;
    }
}
$ y_{vertex} \leftarrow$ \InferVertex{$N_{Root}$, $A$, $X$, $i$}\;

$y_{graph} \leftarrow$ \InferGraph{$N_{Root}$, $A$, $X$}\;
\caption{Infer TREE-G}
\label{alg:infer_treeg}
\end{algorithm}

\acronym preserves the same framework, but instead of using a single feature that will be compared to a threshold as a split function, a more elaborate function $\phi$ is used to integrate vertex features with graph structure. 
All other parts of the decision trees framework are preserved, including the greedy training process. Hence, it can be embedded in ensemble methods, such as Random Forests or Gradient Boosted Trees\footnote{In our experiments, we used GBT with TREE-G estimators} as well as be limited to a certain size either by limiting its depth during training or by using pruning.

The split function used in \acronym incorporates the vertex features, adjacency matrix, and a subset of the vertices of the graph. 
It is parameterized by $5$ parameters $k, d, \pointer, r$ and $\rho$ for vertex labeling tasks, and an additional parameter $AGG$ for graph labeling tasks, as explained in detail in \Secref{subsection:vertex_level} and \Secref{subsection:graph_level_tasks}. 
Overall, while the binary rule in standard trees is of the form $\phi_k(\x) > \theta$, the binary rule in TREE-G is of the form $\phi_{k, d, \pointer, \rho ,r}(X, A, i) > \theta$ for vertex labeling tasks and $\phi_{k, d, \pointer, \rho ,r, AGG }(X, A) > \theta$ for graph labeling tasks. Here $X$ is a matrix of the stacked vertex feature vectors, $A$ is an adjacency matrix, and $i$ is an index of a vertex in the graph.
Algorithm ~\ref{alg:infer_treeg} presents the pseudo-code for the inference process in TREE-G.
Notice that it only differs from the standard DT inference procedure in the split function $\phi$ that is computed and compared against the threshold during the traversal of the tree.

For the rest of this paper, we will focus only on the novel split function $\phi_{k, d, \pointer, \rho ,r}$ TREE-G introduces, and how it enables learning effective decision trees for graph and vertex predictive tasks.
For the sake of clarity, we first focus on vertex labeling, and explain graph labeling in \Secref{subsection:graph_level_tasks}.

\subsubsection{\acronym for Vertex  Labeling}\label{subsection:vertex_level}
We begin with the assumption of a trained tree and demonstrate its use during the inference phase. Training details are provided in \Secref{sec:training}.
In vertex labeling, a vertex $i$ is given for inference, together with its graph $G$ and vertices' feature matrix $X$.
The split function applied to the input in \acronym is based on a feature, say $\f_k$, propagated through walks of length $d$ over parts of the graph. For example, for some vertex $i$ in a graph, the question asked by the split function may be: is the sum of the $k$\textsuperscript{th} feature, limited to the neighbors of the vertex $i$, greater than $\theta$?  Moreover, the decision can use information only from vertices that satisfy additional constraints but we begin the exposition without such constraints and defer this discussion to later.

Splits are based on propagation of feature values over walks in the graph that are computed by the function $A^d\f_k$. Here, the parameters $d$ and $k$ are used to define the propagation of the $k$\textsuperscript{th} feature through walks of length $d$ in the graph. When $d=0$ the graph structure is ignored since $A^0 \f_k = \f_k$. However, when $d=1$ for example, the $i$\textsuperscript{th} entry in $A^1 \f_k$  is the sum of the values of the $k$\textsuperscript{th} feature over the neighbors of the $i$\textsuperscript{th} vertex. In directed graphs it may be useful to consider walks in the opposite direction by taking powers of $A^T$. Note also that if $\f_0$ is the constant feature $1$, then $(A^d \f_0)_i$ is the number of walks
of length $d$ that end in vertex $i$.

Next, we introduce a way to further restrict the walks used for feature propagation.  
We would like to adapt the values of $A^d\f_k$ to focus on a subset of the vertices $\att$.
Hence, the split function can use a graph's substructure rather than its entirety. We present intuitive and computationally efficient methods below for this purpose.\footnote{This selection allows computing only once the powers of the adjacency matrix, as a pre-processing step. See more in the Appendix.}

\begin{enumerate}
\item \textbf{Source Walks}: Only consider walks starting in the vertices in $\att$. 
\item \textbf{Cycle Walks}: Only consider walks starting and ending in the same vertex in $\att$. 
\end{enumerate}

These restrictions can be implemented by zeroing out parts of the walks matrix $A^d$ before propagating the feature values $\f_k$ through it.
For source walks, zero out the $j$\textsuperscript{th} column in $A^d$ for each vertex $j \notin \att$. For cycle walks, zero out all $A^d$ except for the $j$\textsuperscript{th} value on the main diagonal, for each vertex $j \in \att$.
Hence, we define a mask matrix that corresponds to one of the two options described above (i.e., source or cycle walks), and denote it by $M_r(\att), r\in \{1,2\}$. The parameter $r$ indicates the type of walk restrictions, where $r=1$ indicates source walks and $r=2$ indicates cycle walks.  We apply $M_r(\att)$ to $A^d$ using an element-wise multiplication denoted by $\circ$.

The key challenge is to design a policy to select the set $\att$. Clearly, we cannot add $\att$ as a feature that the DT enumerates over, since there are exponentially many and their definition would not generalize across different graphs. Our main insight involves determining the subset $\att$ only after receiving the graph $G$. To achieve that, we utilize ancestor split-nodes to create candidate subsets of vertices to choose from.
To that end, in each split-node in \acronym two actions are performed: (1) the split function is applied to the example to determine to which child it should continue, and (2) the split-node generates two subsets of the vertices of the given graph. We denote the two generated subsets at a split-node $u$ to be $\att_{u,+}$ and $\att_{u,-}$.
For clarity, we'll first present the full expression for the split function $\phi$ and defer the detailed definitions of these subsets to \Secref{subsection:subset_generation}.

Every split-node in the tree uses one of the subsets generated by its ancestors in the tree. In a node $u$, the parameter $\pointer_u$ points to the ancestor and the parameter $\rho_u\in \{+,-\}$ indicates if the chosen subset is $\att_{\pointer_u,+}$ or $\att_{\pointer_u,-}$.
Therefore, the subset $\att_{\pointer_u, \rho_u}$ to be used in node $u$ is well-defined when $u$ has to evaluate the input graph $G$ since all its ancestor nodes have already been evaluated. 
Notice that $\att_{\pointer_u,\rho_u}$ is graph dependent, and therefore translates to different subsets for different graphs. To allow a split-node to focus on the entire graph we use the convention that if $\pointer_u \equiv u$ then $\att_{\pointer_u,\rho_u} = V$, regardless of the value of $\rho_u$. 
Since the root node has no ancestors, the root node always points to itself and therefore uses all the vertices as a subset.


Finally, given a vertex example $i$, the adjacency matrix $A$, and the stacked feature vectors of the graph's vertices $X$, the split function \acronym computes is defined as follows:
\begin{equation}
\phi_{k,d,\pointer,\rho,  r} (A,X,i) =\left(\left(A^d\circ M_r(\att_{\pointer,\rho})\right) \f_k  \right)_i
\end{equation}
Then, the binary decision at a split-node will be given by the binary rule:
\begin{equation}
\phi_{k,d,\pointer,\rho,  r} (A,X,i) > \theta
\label{eq:decision_func_at_node}
\end{equation}

     \begin{figure}[t]
    \centering
    \includegraphics[width=1\linewidth]{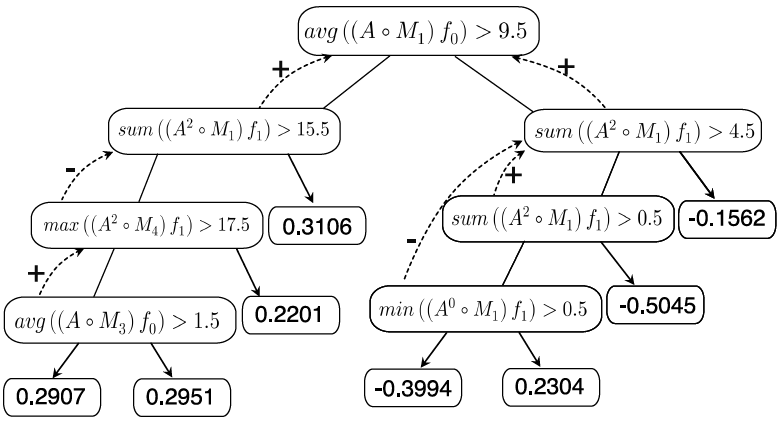}
    \caption{One \acronym tree in the ensemble of the Mutag experiment. Each node in the tree presents the split function and threshold in that node. The dashed arrow in each node is the pointer $\pointer$ and it points to the ancestor split-node where the subset is taken from together with the value of $\rho \in \{+,-\}$ which indicates which subset from the two subsets generated in that ancestor to use.}
    \label{Figure:mutag_tree}
\end{figure}

 \begin{figure*}[t]
    \centering
   \includegraphics[width=1\linewidth]{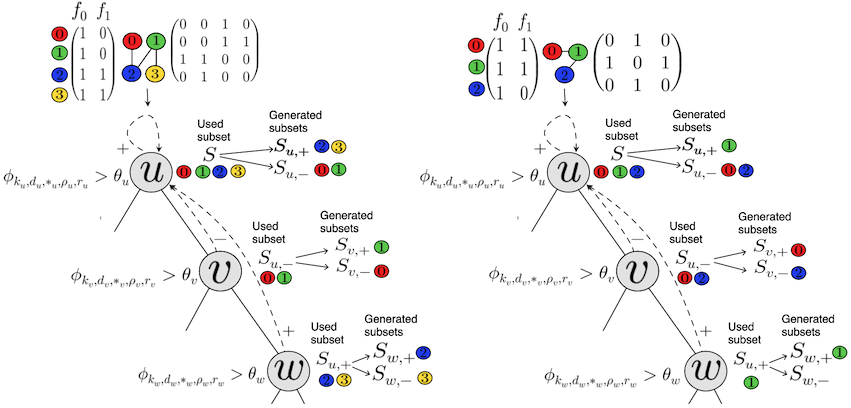}
    \caption{The same trained \acronym instance is applied to two graphs of different sizes during inference. Each split-node uses one subset among the subsets generated in its ancestor nodes, and the set of all vertices $S$. The subset to use in each split-node is uniquely defined by a pointer $*$ that points to the ancestor where the subset is generated, and $\rho\in \{+,-\}$ that indicates which of the two subsets generated in that ancestors should be used. Each split-node generates two subsets, by splitting its used subset.
    The subsets are computed using the same rules but translated to different subsets with respect to the given graphs.
    }
        \label{Figure:attention}
    \end{figure*}

Figure~\ref{Figure:mutag_tree} presents one \acronym tree in an ensemble of trees learned in the evaluation presented in Section~\ref{Section:experiments}, on the Mutag dataset. Each split-node in the tree shows the split function used in that split-node. The dashed arrows in each split-node represent the value of the $\pointer$ parameter and point to an ancestor in the path to the root where the used subset is taken from. The value of the $\rho$ parameter appears on each arrow, and it indicates which of the two generated subsets in that ancestor should be used.

\subsection{Subsets Generation}\label{subsection:subset_generation}
As mentioned above, each split-node $u$ generates two subsets of the vertices $\att_{u,+}$ and $\att_{u,-}$, which may be consumed by its descendant split-nodes in the tree.
Recall that a split function in $u$ uses the parameters $k_u,d_u,\pointer_u, \rho_u, r_u$ and a threshold $\theta_u$. 
Informally speaking $\att_{u,+}$ are the vertices that increase the value of $\phi=\phi_{k_u,d_u,\pointer_u,\rho_u,r_u}$ to be greater than $\theta_u$ while $\att_{u,-}$ are the vertices that decrease $\phi$ to be smaller than $\theta_u$. Formally speaking, 
the node $u$ already uses a subset of the vertices $\att_{\pointer_u, \rho_u}$. 
Then, the generated subset $S_{u,+}$ contains the vertices among $S_{\pointer_u, \rho_u}$ that satisfy the criterion in \Eqref{eq:decision_func_at_node} with the selected parameters and threshold. Similarly, the generated subset $S_{u,-}$ is the vertices among $S_{\pointer_u, \rho_u}$ that do not satisfy the criterion. That it:
\begin{equation}\label{eq:subset_generation}
\begin{aligned}
&S_{u,+} = \{j | \phi(A,X,j) > \theta_u,~~~j\in S_{\pointer_u, \rho_u} \}\\
&S_{u,-} = \{j | \phi(A,X,j) \leq \theta_u,~~~j\in S_{\pointer_u, \rho_u}\}
\end{aligned}
\end{equation}
where $\phi=\phi_{k_u,d_u,\pointer_u,\rho_u,r_u}$.
In particular, it holds that  $S_{u,+} \cap S_{u,-}  = \emptyset$ and $S_{u,+} \cup S_{u,-} = S_{\pointer_u, \rho_u}$.
Notice that subsets are generated separately for every different graph, however, they are generated from the same rule. Moreover, the definition of subsets is invariant to isomorphism and it does not assume a graph size, which allows \acronym to be applied to graphs of varying sizes including sizes that were not seen during training.
Notice that in vertex-labeling tasks, each split-node splits the vertices that arrive at it into two sets, one that would proceed left and the other would proceed right when traversing the tree. At the same time, it also splits the vertices of its used subsets into two sets $\att_{\cdot,+}$ and $\att_{\cdot,-}$. Since the vertices that arrive at a node are not necessarily the vertices composing its used subset, it is also true that $\att_{\cdot,+}$ and $\att_{\cdot,-}$ do not necessarily correspond with the vertices that traverse left and right.

\Figref{Figure:attention} demonstrates how the same trained \acronym instance is applied to two graphs of different sizes during inference. Each split-node uses one subset among the subsets generated in its ancestor nodes, and the set of all vertices $S$. The subset to use in each split-node is uniquely defined by a pointer $*$  that points to the ancestor where the subset is generated (marked by dashed lines), and $\rho\in \{+,-\}$ that indicates which of the two subsets generated in that ancestors should be used. Each split-node generates two subsets, by splitting its used subset. The subsets are computed using the same rules but translate to different subsets with respect to the given graphs.
Since node $u$ is the root, it points to itself and uses the set $S$ of all vertices of the graph. It generates two subsets $S_{u,+}, S_{u, -}$ by partitioning $V$. For both graph examples, the same function is applied to generate $S_{u,+}, S_{u, -}$, but it results in different subsets that correspond to the given graphs.
Split-node $v$ points to its ancestor $u$ with $\rho_v=-$, therefore $\att_{\pointer_v, \rho_v} = \att_{u,-}$. It generates two subsets by partitioning its used subset $S_{u,+}$ into two new subsets: $S_{v,+}$ and $S_{v,-}$.
Node $w$ also points to its ancestor $u$, with $\rho_w=+$. Therefore it uses the subset $S_{u,+}$, and generated two subsets by splitting it.
For the sake of clarity in this figure, we provide the generic split function. Our goal in this figure is to show the general mechanism and how the same instance is applied to different graphs. We provide a detailed example of the split functions that produce these subsets, as well as an example of a trained \acronym over a real-world dataset, in the Appendix.

\Lemref{thm:GTA_ge_dt} shows the theoretical merits of a split function that can focus on subsets. Specifically, it demonstrates that without this mechanism there exist graphs with merely $4$ vertices that are indistinguishable.

The selected subsets can also be used to explain the decisions a TREE-G makes by noting that vertices that appear more often in the used subsets are likely to have a larger influence on the prediction. Therefore, it is possible to visualize these vertices as an explainability aid. Due to space limitations, this is discussed and demonstrated at length only in the Appendix.

\subsection{Training \acronym{}s}\label{sec:training}
As \acronym preserves most of the standard DTs framework, training it uses the same greedy process as standard DTs. First, an optimization criterion is selected, for example, the Gini score or the $L_2$-loss. For each leaf, a grid search is used to find the parameters that will reduce the score the most if the leaf was to become a split-node and the potential score reduction is recorded. The leaf that creates the largest score reduction is selected and is converted into a split-node using the optimal parameters that were found. This process repeats until a stop criterion is satisfied. Popular criteria include bounds on the size of the tree, bounds on the number of training examples that reach each leaf, or bounds on the loss reduction. 
The pseudo-code for the training procedure of TREE-G is given in Algorithm~\ref{alg:train_treeg}.
Note that the only difference between the training process of TREE-G and the training process of standard DTs is in the grid search for the optimal split parameters where TREE-G tunes additional parameters according to the special split function it uses. This induces a penalty in terms of running time that can be controlled (see \Secref{sec:theoretical_analysis}).
For example, it helps to bound the possible values of $d$. Bounding the search space for the pointer $\pointer$ can also improve efficiency.
In particular, it is possible to limit every split-node to consider only pointers to ancestors at a limited distance $a$ from it.
Empirically, we discovered that bounding $d\leq 2$ and $a\leq 2$
is sufficient to achieve high performance without incurring a significant penalty.

In the description above we described the way a single \acronym is learned and used. Multiple \acronym{}s can be used to form ensembles using Gradient Boosted Trees~\cite{gbt}, Random Forests~\cite{rf}, or any other ensemble learning method.

\begin{algorithm}[t]
\SetKwInput{KwPreDefined}{Pre-Defined}
\KwData{$D$: training dataset of labeled graphs or labeled nodes.}
\KwResult{Root of a trained TREE-G $T$}
\KwPreDefined{Optimization criterion (e.g., Gini score or the $L_2$ loss), stop conditions (e.g., minimal gain, maximal depth, minimal examples in leaf), 
label computation rule (e.g., majority vote or average of labels)}
\SetKwFunction{Train}{TrainTREE-G}
\SetKwFunction{SplitLeaf}{SplitLeaf}
\SetKwFunction{CreateLeaf}{CreateLeaf}
\SetKwProg{Fn}{Function}{:}{}

\Fn{\Train{$D$}}{

    $root \leftarrow$ \CreateLeaf($D$)\;
    
    \While {No stop condition is met}{
        N $\leftarrow$ leaf with maximal potential gain with respect to the pre-defined optimization criterion\;
        
        \SplitLeaf($N$)\;        
     }
    \Return root\;
}
\Fn{\CreateLeaf{$D$}}{

    Create a node $N$\;
    
    Compute the label using the label computation rule and store it in $N$\;

    Store the examples $D$ in $N$ as $D_N$\;

    \Return $N$
}

\Fn{\SplitLeaf{$N$}}{

    Select the best split parameters $k_N,d_N,\pointer_N,\rho_N, r_N$, threshold $\theta_N$ and $\agg_N$ in the case of graph labeling, with respect to the pre-defined optimization criterion\;

    For each different graph in $D_N$ generate two subsets of the vertices of $G$, $S_{N,+}, S_{N,-}$ using the chosen parameters of $N$, following Equation~\ref{eq:subset_generation}, and store in $N$.\;
    
    Split the examples $D_N$ into two disjoint subsets $D_1$ and $D_2$ by applying the split function to each example in $D_N$, and comparing it to the threshold: $\phi_{k_N,d_N,\pointer_N,\rho_N, r_N} > \theta_N$ for vertex labeling or $\phi_{k_N,d_N,\pointer_N,\rho_N, r_N, \agg_N} > $ for graph labeling\;
    
    Create two child leafs for $N$:\\
    \CreateLeaf($D_1$) \\
    \CreateLeaf($D_2$)
    
    \Return the split-node $N$\;
}
$T \leftarrow$ \Train{$D$}\;

\caption{Train TREE-G}
\label{alg:train_treeg}
\end{algorithm}

\subsection{\acronym for Graph Labeling}\label{subsection:graph_level_tasks}
So far we have discussed \acronym in the context of vertex labeling tasks. 
We now shift to graph labeling tasks in which a graph is given and its label needs to be predicted. While the general structure of \acronym is preserved, some modifications are required.

In graph labeling tasks, only the adjacency matrix $A$ and the matrix of stacked feature vectors $X$ are given. Since there is no target vertex $i$, instead of taking some entry of the vector $\left(A^d\circ M_r\left(S_{u,\rho}\right)\right)\f_k$ to compare against a threshold, the split function aggregates all its entries to produce a scalar. This aggregation should be permutation-invariant to be stable under graph isomorphism. We use one of $sum, mean, min, max$ for this purpose. The selected aggregation function is specified as another parameter, $\agg$, to the split function $\phi$.
Hence, the split function used in graph labeling tasks is defined as follows:
\begin{equation*}
\phi_{k,d,\pointer,\rho, r, AGG} (A,X) =AGG\left(\left(A^d\circ M_r\left(S_{\pointer,\rho}\right)\right) \f_k  \right)
\end{equation*}

In addition, for graph labeling tasks, two additional walk types are used:
\begin{enumerate}
\setcounter{enumi}{2}
\item \textbf{Target Walks}: Only consider walks ending in the vertices in $\att$.
\item \textbf{Target-Source Walks}:  Only consider walks starting and ending in  vertices of $\att$.
\end{enumerate}
Therefore the parameter $r$ can take a value of $3$ for target walks or $4$ for target-source walks. Similarly to types $1$ and $2$, it can be implemented by masking some entries of the matrix $A^d$ given a subset of the vertices $\att$.
For target walks, zero out are the rows of $A^d$ corresponding to vertices not in $\att$. 
For target-source walks, zero out are the rows and columns of $A^d$ corresponding to vertices not in $\att$. In the Appendix we empirically show that the four walk types are not superfluous.
Notice that target masking (including target-source) is not useful and leads to redundant computation in the vertex labeling setting since in this setting only the $i$\textsuperscript{th} coordinate of the vector $\left(A^d\circ M_r\left(S_{\pointer,\rho}\right)\right)\f_k$ is used when making predictions for the $i$\textsuperscript{th} vertex. However, in the graph labeling, predictions are made for the entire graph and use the entire vector.

Masking types that include zeroing out rows, eventually zero out entries in the vector $\left(A^d\circ M_r\left(S_{\pointer,\rho}\right)\right)\f_k$. As in graph labeling tasks this vector is aggregated, zeros may affect the results of some aggregations e.g. $min$. Therefore, in the case of target, target-source and cycle walks, the aggregation is only performed on the entries that are in the selected subset $S_{\pointer,\rho}$; i.e., $AGG \left(\left(\left( A^d\circ M_r\left(S_{\pointer,\rho}\right)\right) \f_k\right) _{S_{\pointer,\rho}}\right)$

As in the vertex labeling setting, after the graph is introduced to a split-node, two subsets of the vertices are generated to be used by the descendant split-nodes in the tree, using the same definitions as in vertex labeling.
In particular, the chosen aggregation is not part of the subsets generation mechanism. An exception is made when the selected aggregation function in the split-node is \emph{sum}, in which case the subsets are defined with respect to a scaled threshold $\nicefrac{\theta}{\left\vert S_{\pointer,\rho} \right\vert}$ instead of $\theta$. This is because in this case any value that is greater than this scaled threshold contributes to $\phi$ towards passing the threshold.


%% file: theoretical_properties.tex
In this section we discuss some theoretical properties of \acronym, including its computational complexity. Due to space limitations proofs are deferred to the Appendix.
We first consider the invariance properties of \acronym. Recall that a graph is described by its features $\f_1,\f_2,\ldots, \f_l$ and adjacency matrix $A$. If we apply a permutation $\pi$ to the vertices, with the corresponding permutation matrix $P=P_\pi$, we obtain a new adjacency matrix $PAP^T$ and features $P\f_k$. Clearly, we would like a model that acts on graphs to provide the same output for the original and permuted graphs. In graph labeling tasks this means we would like the output of \acronym to be invariant to $P$. 
In the case of vertex labeling, we expect equivariance such that if the graph is permuted with $P$ then the prediction for the vertex $\pi(i)$ would be the same as the prediction for the vertex $i$ in the original graph~\cite{gnns_invariance}. 
The next lemma shows that this is the case for \acronym.

\begin{lemma}\label{lemma:permutationinvariance}
\acronym is invariant to permutations for graph labeling and equivariant for vertex labeling.
\end{lemma}




Next, we discuss the computational complexity of \acronym. The main computational challenge in training decision trees is finding the optimal parameters for the split function. The cost of this operation is proportional to the number of features to consider.
This is also the case in \acronym
\begin{lemma}\label{thm:GTA_running_time}
The running time of searching for the optimal split function parameters in \acronym is linear in: the number of features, the maximal walk length, and the maximal number of considered pointed ancestors.
\end{lemma} 



Next, we present results on the expressive power of \acronym.
As mentioned above, the advantage of \acronym lies in its ability to focus on subsets. We prove that \acronym is strictly more expressive than a limited version of it where subsets are not used (equivalently, the used subset is always $V$).


\begin{lemma}\label{thm:GTA_ge_dt}
There exist graphs that are separable by \acronym , but are inseparable if \acronym is limited to not use subsets.
\end{lemma}

Notice that standard decision trees that only use the input features in split-nodes are a special case of such limited \acronym, where additionally the propagation depth is limited to $0$. Therefore, an immediate conclusion is that \acronym is strictly more expressive than these standard trees. 

Finally, the following lemma shows that \acronym can express classification rules that GNNs cannot.

\begin{lemma}\label{thm:GTA_neq_GNNs}
 There exist graphs that cannot be separated by GNNs but can be separated by \acronym.
\end{lemma}
The Lemma shows that there exist two graphs such that any GNN will assign both graphs with the same label, whereas there exists a \acronym{} that will assign a different label to each graph.
Since the expressive power of GNNs is bounded by the expressive power of the 1-WL test \cite{gin, gnnlimitations} we conclude that the expressive power of \acronym is not bounded by the 1-WL test.

%% file: experiments.tex
\input{tables/split_graph_level_big}
\input{tables/node_classification_res}
\input{tables/split_graph_level_small}

We conducted experiments using \acronym on graph and vertex labeling benchmarks and compared its performance to popular GNNs, graph kernels, and other tree-based methods.\footnote{Code is available at github.com/mayabechlerspeicher/TREE-G}

\subsection{Datasets}
\textbf{Graph Prediction Tasks:}
We used nine graph classification benchmarks from TUDatasets~\cite{tudataset}, and one large-scale dataset from the Open Graph Benchmark~\cite{ogb}.
\emph{IMDb-B \& IMDb-M}~\cite{dgk} are movie collaboration datasets, where each graph is labeled with a genre.
\emph{Mutag, NCI1, Protiens, D\&D , Enzymes, Mutagenicity \& PTC} ~\cite{wl, ptcmr, mutagenicity_dataset} are datasets of chemical compounds. In each dataset, the goal is to classify compounds according to some property of interest.
\emph{molHIV}~\cite{ogb} is a large-scale dataset of molecules that may inhibit HIV.\\
\textbf{Node Prediction Tasks:}
We used the three citation graphs tasks \emph{Cora, Citeseer \& Pubmed} from Planetoid \cite{planetoid}. Arxiv is a large scale citation network related task of Computer Science arXiv papers \cite{ogb}. 

\emph{Cornell, Actor \& County} are heterophilic datasets where 
\emph{Cornell} and \emph{Actor}  \cite{Pei2020GeomGCN} are web links networks with the task of classifying nodes to one of five categories whereas the \emph{County} dataset~\cite{jia2020residual} contains a regression task of predicting unemployment rates based on county-level election map network.

More details on all datasets are given in the Appendix.

\subsection{Baselines and Protocols}
We compared \acronym as a weak-learner in GBT to the following popular GNNs and graph-kernels:
Graph Convolution Network (GCN) \cite{gcn}, Graph Isomorphism Network (GIN) \cite{gin}, Graph Attention Network (GAT) \cite{gat} and its newer version GATv2~\cite{gatv2}, GraphSAGE \cite{graphsage},  Weisfeiler-Lehman Graph kernels (WL) \cite{wl} and a Random-Walk kernels (RW) \cite{rw}. We also compared to two methods combining GNNs and DTs: BGNN \cite{ivanov2021boost} and XGraphBoost \cite{xgraphboost}.
Additionally, we report two ablations of \acronym 
where we do not use the topology of the graph and disable the subsets mechanism. This is done by limiting the propagation depth to $0$ and/or the distance of the ancestors from which to consider subsets from, to $0$. In \acronym~($a=0$) we always use the whole graph as the selected subset, which is equivalent to not using subsets. In \acronym~($d=0,a=0$) we do not use the topology of the graph (i.e. we always use $A^0$) and we always use the entire graph as the selected subset. Notice \acronym~($d=0,a=0$) is equivalent to standard trees that only act on the input features.

For graph prediction tasks we report the average accuracy and std of a $10$-fold nested cross-validationn, except for the molHIV dataset for which we use official pre-defined splits provided in \citet{ogb} and the metric is AUC. For the node prediction tasks we report average accuracy and std using the pre-defined splits provided in the data, except for the county dataset, which is a regression task, and hence we use RMSE instead of accuracy.
More details on the evaluation protocol including the tuned hyper-parameters for each algorithm are provided in the Appendix.

\subsection{Results}\label{sec:results}

The results are summarized in Tables \ref{Table:graph_exp_res_big}, \ref{Table:node_exp_res} \& \ref{Table:graph_exp_res_small}.
\acronym outperformed all other tree-based approaches (XGraphBoost, BGNN, \acronym($a=0$), on all $17$ tasks. It outperformed graph kernels in $9$ out of the $10$ graph classification tasks and outperformed GNNs in $7$ out of these $10$ tasks.
Specifically, \acronym improved upon GNN approaches by a margin greater than $6.4$ percentage points on the IMDb-M dataset and a margin greater than $4.9$ and $2.4$ percentage points on Mutag and Proteins datasets, respectively.
On the vertex labeling tasks, \acronym outperformed GNNs in $4$ out of the $7$ tasks and was on par with the leading GNN approach in every task.
Note that \acronym always outperformed \acronym~($a=0$), which indicates that the use of subsets leads to improved performance. 
Consequently, these results demonstrate that a "pure" decision tree method, when tailored for graphs, can rival popular GNNs and graph kernels, potentially even surpassing them.

%% file: tables/split_graph_level_big.tex
\begin{table*}[t]
  \centering\fontsize{10}{11}\selectfont
\begin{tabular}{lccccccc}
Baseline & Proteins & Mutag & D\&D & NCI1 & PTC & Enzymes & Mutagenicity \\
\toprule
XGraphBoost & 67.0 $\pm$ 2.5 & 86.1 $\pm$  3.8 & 72.9 $\pm$  2.0 & 61.9 $\pm$  7.1 & 51.3 $\pm$  5.0 & 58.5 $\pm$ 1.0 & 71.2 $\pm$ 2.5 \\ 
BGNN & 70.5 $\pm$ 2.4 & 80.2 $\pm$ 0.5 & 71.2 $\pm$ 0.9 & 70.5 $\pm$ 2.0 & 55.5 $\pm$ 0.3 & 58.1 $\pm$  1.5 & 65.0 $\pm$  0.9 \\
RW & 72.5 $\pm$ 0.9 & 85.0 $\pm$ 4.5 & 70.0 $\pm$ 5.0 & 69.0 $\pm$ 0.5 & 57.2 $\pm$ 1.0 & 55.1 $\pm$ 1.3 & 74.5 $\pm$ 0.5 \\ 
WL & 74.0 $\pm$ 3.0 & 82.0 $\pm$ 0.2 & 76.1 $\pm$ 3.0 & \textbf{82.5 $\pm$ 3.1} & 58.9 $\pm$ 4.2 & 52.1 $\pm$  3.0 & 80.1 $\pm$  2.0 \\ 
GAT & 70.0 $\pm$ 1.9 & 84.4 $\pm$ 0.6 & 74.4 $\pm$ 0.3 & 74.9 $\pm$ 0.1 & 56.2 $\pm$ 7.3 & 58.5 $\pm$ 1.2 & 72.2 $\pm$ 1.8 \\ 
GCN & 73.2 $\pm$ 1.0 & 84.6 $\pm$ 0.7 & 71.2 $\pm$  2.0 & 76.0 $\pm$  0.9 & 59.4$\pm$ 10.3 & 60.1 $\pm$ 3.0 & 69.9 $\pm$  1.5 \\
GraphSAGE & 70.4 $\pm$ 2.0 & 86.1 $\pm$ 0.7 & 72.7 $\pm$ 1.0 & 76.1 $\pm$  2.2 & \textbf{67.1 $\pm$ 12.6} & 58.2 $\pm$ 5.9 & 64.1 $\pm$  0.3 \\
GIN & 72.2 $\pm$ 1.9 & 84.5 $\pm$ 0.8 & 75.2 $\pm$  2.9 & 79.1 $\pm$  1.4 & 55.6 $\pm$  11.1 & 59.5 $\pm$ 0.2 & 69.4 $\pm$  1.2 \\
GATv2 & 73.5 $\pm$ 0.9 & 84.0 $\pm$ 1.5 & 75.0 $\pm$ 1.3 & 77.0 $\pm$ 0.8 & 59.5 $\pm$ 2.1 & \textbf{61.2 $\pm$ 3.0} & 72.0 $\pm$ 0.9\\
\midrule
\acronym(d=0, a=0) & 73.4  $\pm$  0.9 & 86.1   $\pm$ 6.0 & 74.4 $\pm$ 4.7 & 70.4 $\pm$ 0.7 & 54.5 $\pm$ 1.3 & 58.5 $\pm$ 0.9 & 77.0 $\pm$  1.9 \\
\acronym (a=0) & 74.7  $\pm$  0.9 & 89.3 $\pm$ 7.6 & 76.0 $\pm$ 4.5 & 73.6 $\pm$  2.2 & 59.0 $\pm$ 0.8 & 58.1 $\pm$ 3.9 & 77.2 $\pm$ 1.5 \\
\bottomrule
\textbf{\acronym} & \textbf{75.6  $\pm$ 1.1} & \textbf{91.1 $\pm$ 3.5} & \textbf{76.2 $\pm$ 4.5} & 75.9 $\pm$ 1.5 & 59.1 $\pm$ 0.7 & 59.6 $\pm$  0.9 & \textbf{83.0 $\pm$ 1.7} \\
\end{tabular}

\caption{Empirical comparison of graph classification tasks. For all the datasets the accuracy and std are reported. The best score is highlighted in bold.
}
\label{Table:graph_exp_res_big}
\end{table*}

%% file: tables/node_classification_res.tex
\begin{table*}[t]
  \centering\fontsize{10}{11}\selectfont

\begin{tabular}{lccccccc}
\\
Baseline     & Cora                                & Citeseer                                   &Pubmed                                      & Arxiv &   Cornell & Actor & Country
\\
    \toprule
           XGraphBoost &  60.5 $\pm$ 3.0 & 50.5 $\pm$  3.4 & 61.9 $\pm$  2.2 & 65.8 $\pm$  1.9 & 62.1 $\pm$ 0.3 & 27.2 $\pm$ 1.4 & 3.0 $\pm$ 0.5\\
   BGNN & 71.2 $\pm$ 5.0 & 69.1 $\pm$ 3.9 & 59.9 $\pm$ 0.1 &  67.0 $\pm$ 0.1 & 68.2 $\pm$ 0.5 &  31.1 $\pm$ 0.7 & 1.1 $\pm$ 0.2\\
GAT           & 83.0 $\pm$ 0.6 & 70.0 $\pm$ 1.5 & 79.1 $\pm$ 0.7 & 73.6 $\pm$ 0.1 &  \textbf{74.0 $\pm$ 0.5} & 34.0 $\pm$ 0.9 & 1.6 $\pm$ 0.2 \\
GCN      & 81.0 $\pm$ 0.8 & 72.1 $\pm$ 0.3 & 79.0 $\pm$ 0.8 & 71.7 $\pm$ 0.2  &  65.6 $\pm$ 0.1 & 28.8 $\pm$ 0.2 & 1.8 $\pm$ 0.1\\
GraphSAGE       & 81.4 $\pm$ 0.7 & 73.4 $\pm$ 0.8 & 78.4 $\pm$ 0.4 & 71.7 $\pm$ 0.2  & 71.1 $\pm$ 1.0 & 32.1 $\pm$ 1.5 & 2.0 $\pm$ 0.4\\
GIN    & 80.0 $\pm$ 1.2 & \textbf{75.1 $\pm$ 1.9}& 75.3 $\pm$0.9 & 73.8 $\pm$ 1.4  & 69.0 $\pm$ 1.3 & 30.4 $\pm$ 0.3 & \textbf{0.9 $\pm$ 0.1}  \\
GATv2 & 83.1 $\pm$ 0.9 & 73.9 $\pm$ 1.5 & \textbf{79.4 $\pm$ 0.5} &  74.0 $\pm$ 2.1 & 72.5 $\pm$ 0.7 & 34.2 $\pm$ 1.9 & 1.6 $\pm$ 0.5\\

    \midrule
 \acronym(d=0, a=0)     & 80.3 $\pm$ 2.1 & 70.2 $\pm$ 2.0  & 74.0 $\pm$ 0.5 & 67.1 $\pm$ 1.5  &71.0  $\pm$ 0.9 & 29.1  $\pm$ 0.5 & 1.9 $\pm$ 0.2 \\      
   \acronym (a=0)  &80.8 $\pm$ 0.5 & 74.3 $\pm$ 1.9 & 72.9 $\pm$ 1.6 & 67.5 $\pm$ 0.6      & 71.8   $\pm$ 1.4 & 32.5  $\pm$ 1.7 & 1.2 $\pm$ 0.8    \\  
    \bottomrule
\textbf{\acronym} & \textbf{83.5 $\pm$ 1.5} & 74.5 $\pm$ 1.0 &78.0 $\pm$ 3.1 &\textbf{74.7 $\pm$ 1.1}  & 73.0 $\pm$ 1.4 & \textbf{37.0 $\pm$ 0.4} & \textbf{0.9 $\pm$ 0.1}\\
\end{tabular}
 \caption{Empirical comparison of node labeling tasks. For the country task, the RMSE and std are reported. For the rest of the datasets,  the accuracy and std are reported. The best score is highlighted in bold.
    }
    \label{Table:node_exp_res}
\end{table*}

%% file: tables/split_graph_level_small.tex
\begin{table}[ht]
  \centering\fontsize{9}{11}\selectfont
\begin{tabular}{lccc}

Baseline & IMDb-B & IMDb-M & molHIV \\
\toprule
XGraphBoost & 56.1 $\pm$ 1.4 & 41.5 $\pm$ 0.9 & 60.1 $\pm$ 5.3 \\ 
BGNN & 68.0 $\pm$ 1.5 & 46.2 $\pm$  2.3 & 77.0 $\pm$ 6.1 \\ 
RW & 65.0 $\pm$ 1.2 & 45.1 $\pm$ 0.9 & 76.7 $\pm$ 2.1 \\
WL & 72.8 $\pm$ 1.9 & 50.9 $\pm$ 3.8 & 75.7 $\pm$ 0.8 \\ 
GAT & 70.6 $\pm$ 3.2 & 47.0 $\pm$  1.9 & 82.4 $\pm$ 3.6 \\ 
GCN & 69.4 $\pm$  1.9 & 50.0 $\pm$  3.0 & 76.6 $\pm$ 0.0 \\ 
GraphSAGE & 69.9 $\pm$ 4.4 & 47.8 $\pm$  1.9 & 79.2 $\pm$ 1.2 \\
GIN & 71.2 $\pm$  3.4 & 48.8 $\pm$  5.0 & 77.8 $\pm$ 1.3 \\
GATv2 &  71.5 $\pm$ 1.2 & 47.6 $\pm$ 0.5 & 81.9 $\pm$ 4.5 \\
\midrule
\acronym(d=0, a=0) & 72.6 $\pm$ 2.4 & 52.3 $\pm$ 1.5 & 70.1 $\pm$  0.5 \\ 
\acronym (a=0) & 72.5 $\pm$ 2.7 & 52.7 $\pm$ 1.5 & 78.5 $\pm$ 1.1 \\ 
\bottomrule
\textbf{\acronym} & \textbf{73.0 $\pm$ 2.4} & \textbf{56.4 $\pm$ 0.9} &  \textbf{83.5 $\pm$ 2.9} \\ 
\end{tabular}

\caption{Empirical comparison of graph classification tasks. The AUC and std are reported for the molHIV dataset, and accuracy and std for the rest. The best result is in bold.
}
\label{Table:graph_exp_res_small}
\end{table}

%% file: explanations.tex
In this section, we formalize the explanation mechanism of \acronym.
We show that the selected subsets can be used to explain the predictions made by \acronym. This explanation mechanism ranks the vertices and edges of the graph according to their presence in chosen subsets and thus highlights the parts of the graph that contribute the most to the prediction.

Following \cite{settrees}, the explanation mechanism ranks the vertices of the graph according to their presence in selected subsets and thus highlights the parts of the graph that contribute the most to the prediction.
 As is the case in most tree methods, \acronym does not combine multiple features in the splits, which makes them more interpretable and less prone to overfitting.

    \begin{figure*}[t]
        \centering
        \includegraphics[width=0.3\textwidth]{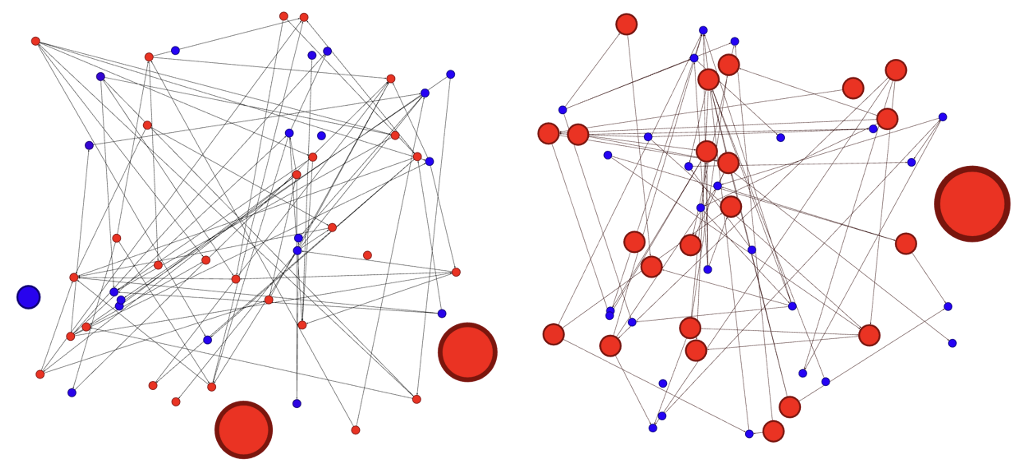}{(a)}
        \includegraphics[width=0.3\textwidth]{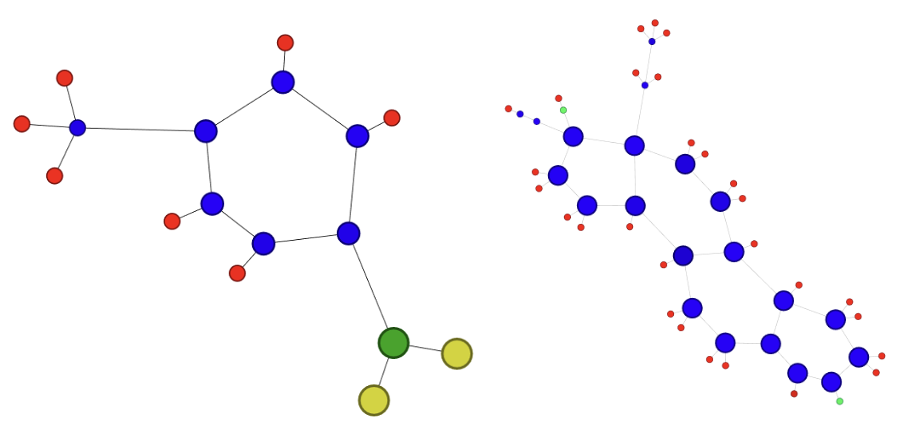}{(b)}

        \caption{Vertex-level explanations for two graphs from the Red Isolated Vertex problem (a) and two graphs from the Mutagenicity problem (b). The size of vertices corresponds to their importance according to the explanation mechanism.} 

        \label{Figure:explanations_example}
    \end{figure*}

Let $G$ be a graph that we want to classify using a decision tree $T$. 
The key idea in the construction of our importance weight is that if a vertex appears in many selected subsets 
sets when calculating the tree output for $G$, then it has a large effect on the decision. This is quantified as follows: consider the path in $T$ that $G$ traverses when it is being classified.
For each vertex $i$ in $G$, count how many selected subsets in the path contain $i$ contain it, and denote this number by $n(i)$. In order to avoid dependence on the size of $T$ we convert the values $n_T(1),\ldots,n_T(|V|)$ into a new vector $r_T(1),\ldots,r_T(|V|)$ where $r_T(i)$ specifies the rank of $n_T(i)$ in $n_T(1),\ldots,n_T(|V|)$ after sorting in decreasing order. For example, $r_T(i)=1$ if the vertex $i$ appears most frequently in chosen subsets compared to any other vertex. We shall use $2^{-r_T(i)}$ to measure the importance of the rank so that low ranks have high importance.

Now assume we have an ensemble of decision trees $\mathcal{T}$ (e.g., generated by GBT). For any $T\in\mathcal{T}$ in the ensemble let $y_T$ be the value predicted by $T$ for the graph $G$.
We weight the rank importance by $\vert y_T \vert$ so that trees with a larger contribution to the decision have a greater effect on the importance measure. Note that boosted trees tend to have large variability in leaf values between trees in the ensemble~\cite{rashmi2015dart}. Therefore, the importance-score of vertex $i$ in the ensemble is the weighted average:   
\begin{equation}{\label{Equations:node_importance}}
    \textit{importance}(i)=\frac{\sum_{T\in\mathcal{T}}|y_T|2^{-r_T(i)}}{\sum_{j\in |V|}\sum_{T\in\mathcal{T}}|y_T|2^{-r_T(j)}}~~~.
\end{equation}
The proposed importance score is non-negative and sums to $1$.
We present examples of vertex explanations for two problems: Red Isolated Vertex, and Mutagenicity.

\paragraph{Red Isolated Vertex} This is a synthetic task to determine whether exactly one red vertex is isolated in the graph. The data consists of $1000$ graphs with $50$ vertices each, with two features: the constant feature $1$ and a binary blue/red color.

\paragraph{Mutagenicity} This is a dataset of molecules where the goal is to classify chemical compounds into mutagen and non-mutagen.

For each task, we trained and tested GBT with $50$ \acronym estimators. We limited the propagation depth to $2$ and the distance of the ancestors from which to consider subsets, to $2$ as well. We used a $80/20$ random train-test splits.  Figures \ref{Figure:explanations_example}a and \ref{Figure:explanations_example}b present vertex explanations for two graphs from the test sets of the Red-Isolated-Node experiments and the Mutagenicity experiments, respectively. The size of a vertex corresponds to its importance computed by the explanations mechanism. For the Red-Isolated-Node problem, it is shown that \acronym attended to isolated vertices, and even more so to red isolated vertices.

For the Mutagenicity problem, \acronym attended to $NO_2$ (green and yellow subgraph) and carbon-rings (blue cycles), which are known to have mutagenic effects~\cite{mutagenicaffect}.

\paragraph{Edge-Level Explanations}{\label{Appendix:edge_level_expla}}
Similarly to vertex importance, edge importance can be computed. We use the selected subsets to rank the edges of a given graph by their \textit{importance}. Note that each split criterion uses a subset of the graph's edges, according to $d$ and the active subset. For example, a split criterion may consider only walks of length $2$ to a specific subset of vertices, which induces the \textit{used} edges by the split criterion. Let $E_1,\ldots,E_k$ denote the used edges along the prediction walk $\mbox{walk}_m(G)$. We now rank each edge $e_{(i,j)}\in E$ by $r_m((i,j))=|\{ j | (i,j) \in E_j\}|$, i.e.,  the rank of an edge is the number of split-nodes in the prediction walk that use this edge. The ranks $r_m(i,j))$ are then combined with the predicted value $y_m$.

%% file: conclusion.tex
Graphs arise in many important applications of machine learning. In this work we proposed a novel method, \acronym, to perform different prediction tasks on graphs, inspired by the success of tree-based methods on tabular data. Our empirical results revealed that \acronym surpassed other tree-based models. Additionally, when compared with prevalent GNNs and graph kernels, \acronym matched their performance in all scenarios and exceeded them in several benchmarks. Our theoretical analysis showed that allowing trees to focus on substructures of graphs strictly improves their expressive power. 
We also showed that the expressive power of \acronym is different from the expressive power of GNNs, as there are graphs that GNNs fail to tell apart, while \acronym manages to separate them. Finally, we also proposed visualizations for \acronym and the predictions in makes to support explainability.

%% file: appendix.tex
\section{Lemmas' Proofs}
The proofs of the lemmas presented in the main paper were skipped due to space limitations. Here we provide the proofs for these lemmas.
\subsection{Proof of Lemma 4.1}{\label{Appendix:prooftlemma4.1}}
\input{invariance_lemma_proof.tex}

\subsection{Proof of Lemma
4.2}{\label{Appendix:prooflemma4.2}}
\input{gta_running_time_lemma_proof.tex}


\subsection{Proof of Lemma
4.3}{\label{Appendix:prooflemma4.3}}
\input{gta_geq_dt_lemma_proof.tex}

\subsection{Proof of Lemma
4.4}{\label{Appendix:prooflemma4.4}}
\input{gta_neq_gnns_lemma_proof.tex}

\section{\acronym Examples and Extensions}
In this section we propose some possible extensions to the \acronym.

\subsection{Weighted Adjacencies and Multiple Graphs}
We note that the adjacency matrix used in \acronym can be weighted. Moreover, it is possible to extend \acronym to allow multiple adjacency matrices. This allows, for example,  the encoding of edge features and heterogeneous graphs.

\subsection{Edge Level Tasks}\label{Appendix:edge_level_tasks}
In the paper, we discuss vertex and graph-level tasks. Nevertheless, edge-level tasks also arise in many problems.

We note that the task of edge labeling can be represented as a vertex-level task by using a line graph~\cite{linegraphs}. The line graph $L(G)$ of a graph $G=(V,E)$ represents the adjacencies between the edges of $G$. Each edge $(u,v)\in E$ corresponds to a vertex in $L(G)$, and two vertices in $L(G)$ are connected by an edge if their corresponding edges in $G$ share a vertex. Then, \acronym for vertex-level tasks can be used to label vertices in the line graph, which corresponds to edges in the original graph.



\section{Additional Experiments Details}

\subsection{Datasets Summary}\label{Appendix:data_summary}
\input{tables/dataset_sum}
Table~\ref{Table:data_sum} presents a summary of the datasets used in the experiments. The datasets used vary in the number of graphs, the sizes of the graphs, the number of features in each vertex, and the number of classes.\\
\textbf{Mutag}~\cite{mutag} is a dataset of 
$188$ chemical compounds divided into two classes according to their mutagenic effect on bacterium.\\
\textbf{IMDb-B, IMDb-M}~\cite{dgk} are  movie collaboration datasets with $1000$ and $1500$ graphs respectively. Each graph is derived from a genre, and the task is to predict this genre from the graph. Nodes represent actors/actresses and edges connect them if they have appeared in the same movie.\\
\textbf{Proteins, D\&D, Enzymes}~\cite{wl, proteins} are datasets of chemical compounds consisting of $1113$ and $1178$ and $600$ graphs, respectively. The goal in the first two datasets is to predict whether a compound is an enzyme or not, and the goal in the last datasets is to classify the type of an enzyme among $6$ classes.\\
\textbf{NCI1}~\cite{wl} is a datasets consisting of $4110$ graphs, representing chemical compounds. Vertices and edges represent atoms and the chemical bonds between them. The graphs are divided into two classes according to their ability to suppress or inhibit tumor growth.\\
\textbf{Mutagenicity}~\cite{mutagenicity_dataset} is a dataset consisting of $4337$ chemical compounds of drugs divided into two classes: mutagen and non-mutagen. A mutagen is a compound that changes genetic material such as DNA, and increases mutation frequency.\\
\textbf{PTC}~\cite{ptcmr} is a dataset consisting of $344$ chemical compounds divided into two classes according to their carcinogenicity for rats.\\
\textbf{molHIV}~\cite{ogb} is a large-scale dataset consisting of $166k$ molecules and the task is to predict whether a molecule inhibits HIV.\\
\textbf{Cornell}~\cite{10.5555/295240.295725} is a heterophilic webpage dataset collected from the computer science department at Cornell University.
Nodes represent web pages, and edges are hyperlinks between them. The task is to classify the nodes
into one of five categories.\\
\textbf{Actor}~\cite{Pei2020GeomGCN} is a heterophilic dataset where nodes correspond to an actor, and the edge between
two nodes denote co-occurrence on the same Wikipedia page.
The task is to classify each actor into one of five categories.\\
\textbf{Country}~\cite{ivanov2021boost} is a county-level election map network. Each node is a
county, and two nodes are connected if they share a border. The task is to predict the unemployment rate.

\subsection{Implementation Details}
The implementation of \acronym is available in the code appendix. For the experiments, we used StarBoost for the GBDT framework.
For the GNNs we used the models from Pytorch Geometric~\cite{pyg}. For the Graph Kernels we used the models from Grakel~\cite{JMLR:v21:18-370}. For the BGNN we used the implementation from~\citet{ivanov2021boost}. For the Xgraphboost we used the implementation from~\citet{xgraphboost}, with GCN as given in the implementation.
For multi-class tasks, we train \acronym  in a one-vs-all approach.
We used NVIDIA GPU Quadro RTX 8000 for network training.
\acronym runs on CPU. We used Sparse Matrix Multiplication to improve efficiency on large graphs.
The random seed generators are available in the code appendix.

\subsection{Protocols}
\paragraph{OGB Datasets Protocols}
The molHIV and ARXIV datasets are large-scale datasets provided in the Open Graph Benchmark (OGB) paper~\cite{ogb} with pre-defined train and test splits and different metrics and protocols for each dataset. As common in the literature when evaluating OGB datasets, for each of these two datasets we follow its defined metric and protocol:
\begin{itemize}
    \item In the case of the HIV dataset, the used metric is ROC-AUC. As described in \citet{ogb}, we ran the \acronym $10$ times and reported the mean ROC-AUC and std over the runs.
    \item In the case of the ARXIV dataset, the metric used is accuracy. Following the protocol in ~\citet{ogb}, we ran \acronym  $10$ times and reported the mean accuracy and std over the runs.
\end{itemize}

\paragraph{Heterophilic Datasets Protocols}
For the Cornell and Actor datasets we used the splits  and protocol from~\cite{Pei2020GeomGCN} and report the test accuracy averaged over $10$ runs, using the best hyper-paremeters found on the validation set. For the Country dataset we used the splits and protocol from~\cite{ivanov2021boost} and report the RMSE averaged over $5$ runs, using the best hyper-paremeters found on the validation set.

\paragraph{Citation Networks Datasets Protocols}
Following~\cite{gcn, graphsage, gat}, for the Core, Citeseer and Pubmed datasets we tuned the parameters on the Cora dataset using the pre-defined splits from~\cite{gcn}. For all these datasets we report the test accuracies averaged over $10$ runs, using the parameters obtained from the best accuracy on the validation set of Cora.

\paragraph{TU-Datasets Protocols}
The PROT, MUTAG, DD, NCI1, PTC, IMDB-B, IMDB-M, Mutagen, and ENZ were trained and evaluated using a $10$-fold nested cross-validation, with $5$ inner fords.
The final reported test accuracies are averages over the $10$ test sets accuracy of the outer $10$ folds.

\subsection{Hyper-Parameters}

\paragraph{\acronym}
\acronym uses a GBDT framework with $\{20, 50\}$ estimators, learning rate of $0.1$. We used max depth $10$, max walk length $d\in\{0, 1, 2\}$ and max ancestor distance $a\in\{0, 1, 2\}$ and $p=0.25$ for graph-level tasks and $p=0.5$ for vertex level tasks.

\paragraph{GNNs} All GNNs use ReLU activations with  $\{2,3,5\}$ layers and $\{32, 64\}$ hidden channels. They were trained with Adam optimizer over $1000$ epochs and early on the validation loss with a patient of steps $100$. We used Weight Decay of $1e-4$ and learning rate in $\{1e-3, 1e-4$\}. We used dropout rate $\{0, 0.5\}$.

\paragraph{Graph Kernels}
The RW kernel uses the fast computation of graph kernels~\cite{10.5555/2976456.2976638} with $\{geometric, exponential\}$ kernel types, and $\lambda \in \{0.1, 0.01\}$, $p\in \{None, 100\}$.
The WL kernel uses ${0-8}$ iterations, following~\citet{Borgwardt_2020}.

\paragraph{XGraphBoost}
We followed the hyper-parameters in~\citet{xgraphboost} used for the tree with learning rate in $\{0.1, 0.01\}$, max depth $4$
min child weight $10$ and $\in\{50, 100\}$ estimators. The GCN feature extractor in~\citet{xgraphboost} uses $\{3, 5\}$ layers and hidden channels in $\{32, 64\}$. They were trained with Adam optimizer over $1000$ epochs and early on the validation loss with a patient of steps $100$. We used Weight Decay of $1e-4$ and learning rate in $\{1e-3, 1e-4$\}.

\paragraph{BGNN}
We followed~\cite{ivanov2021boost} and the number of trees and backward passes per epoch is $\{10, 20\}$, and the depth of the tree is $6$.
The GNNs are among $\{gcn, gat\}$ with $\{2, 3\}$ layers and $64$ hidden dimension. Trained over $1000$ epochs and early on the validation loss with a patient of steps $100$ with Adam optimizer. We used a dropout rate of $\{0, 0.5\}$, weight decay of $1e-4$ and learning rate in $\{1e-2, 1e-3\}$.

\section{Additional Experiments}
\begin{figure}[h]
        \centering
        \includegraphics[width=0.45\textwidth]{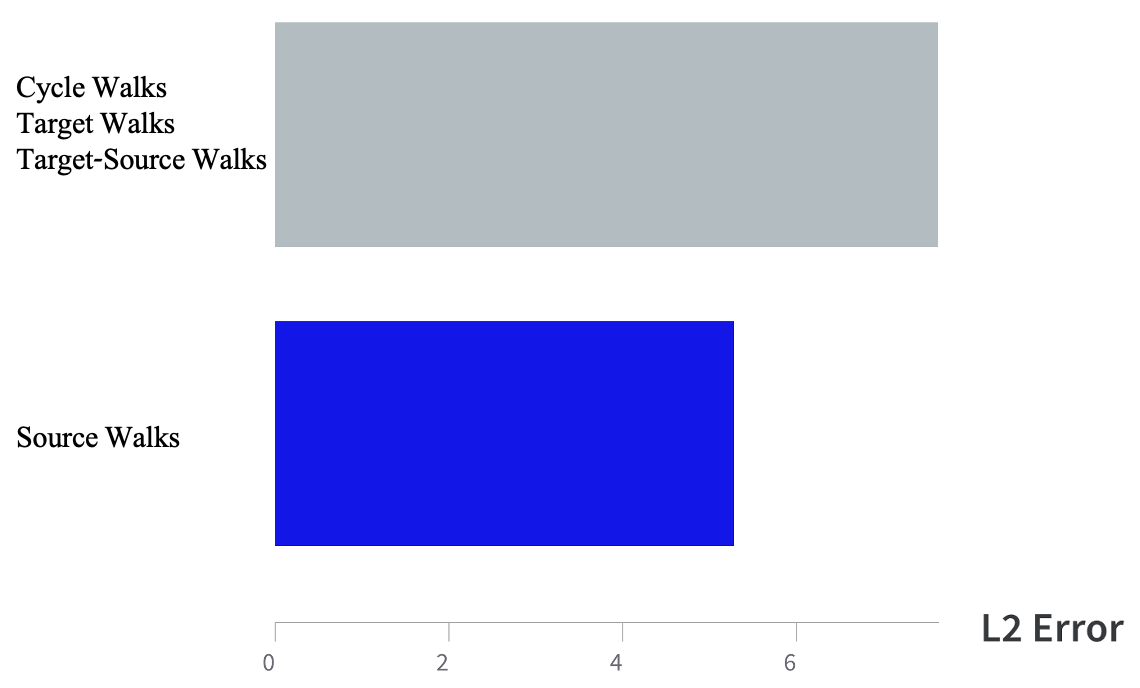}
        \includegraphics[width=0.45\textwidth]{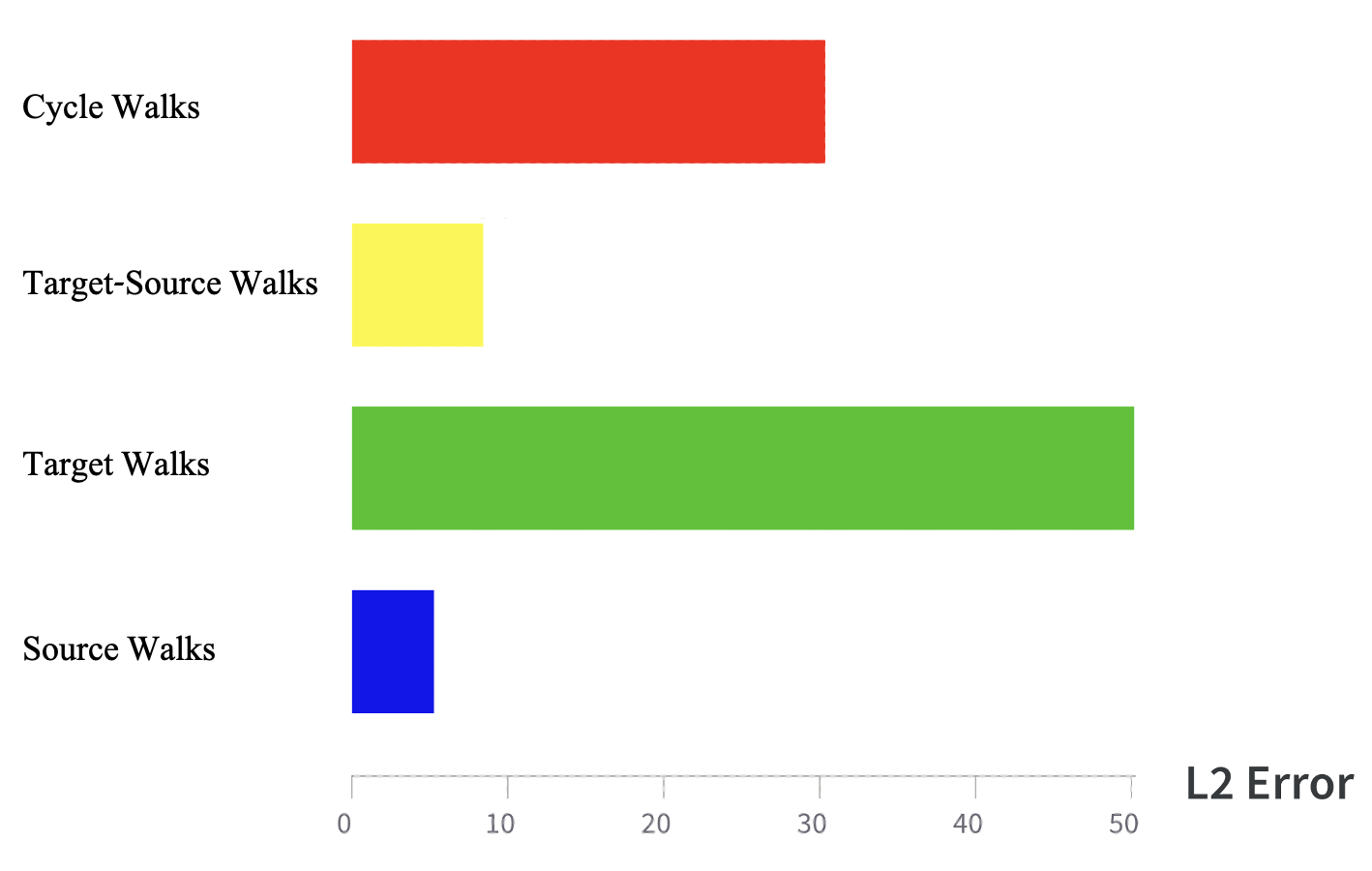}

        \caption{The L2 error of \acronym on the "count walks starting in red vertices" task, tested with different walk types.} 

        \label{Figure:attention_type_1_selection}
    \end{figure}

\begin{figure}[h]
        \centering
        \includegraphics[width=0.45\textwidth]{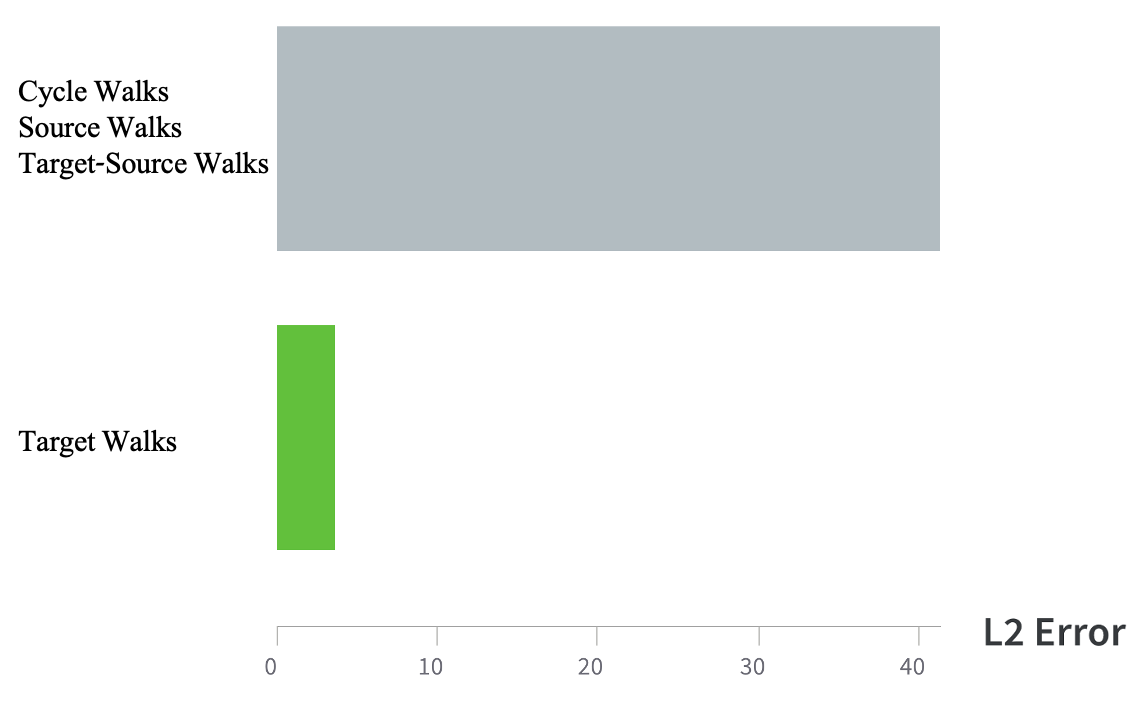}
        \includegraphics[width=0.45\textwidth]{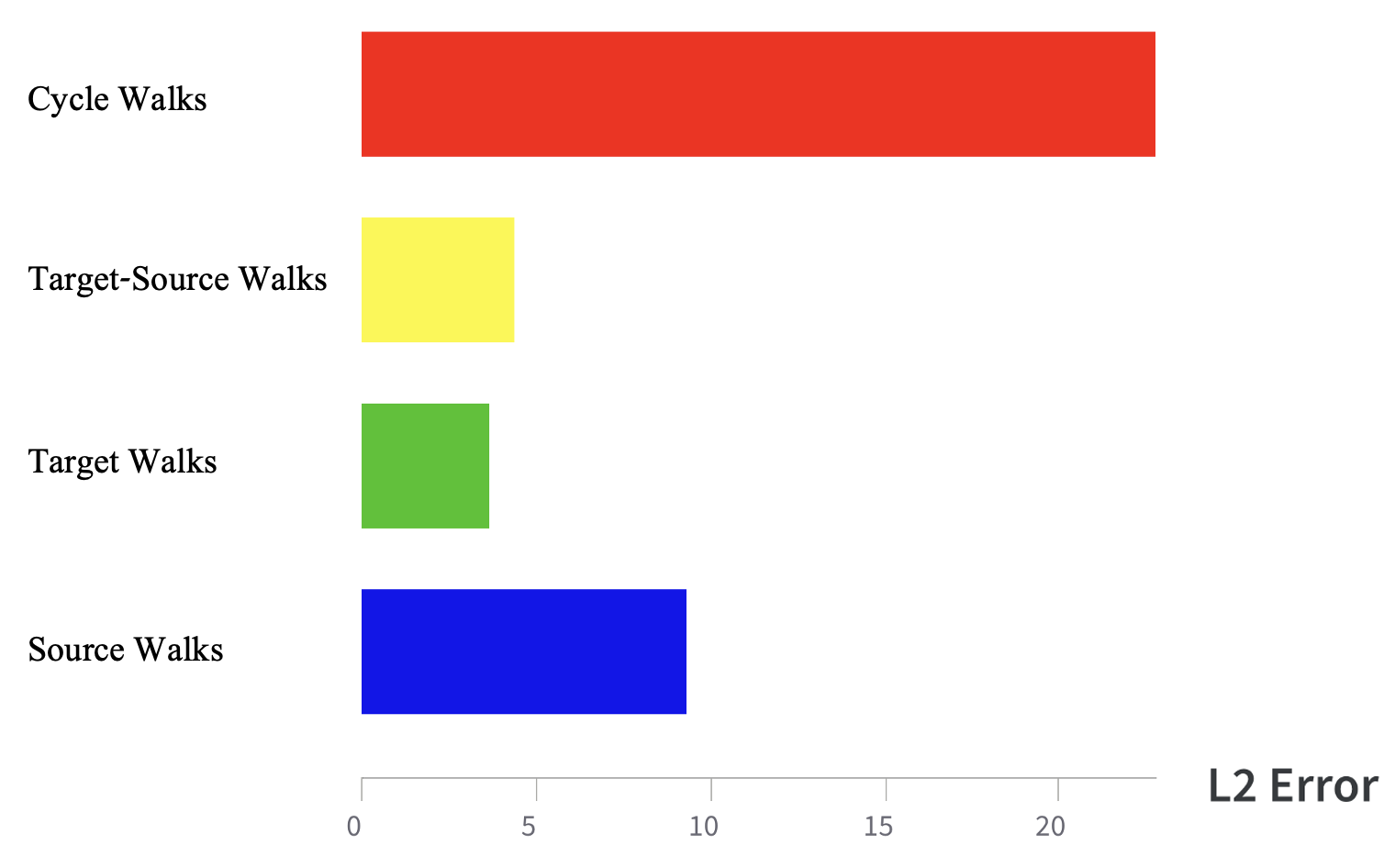}

        \caption{The L2 error of \acronym on the "Count walks ending in red vertices" task, tested with different walk types.} 

        \label{Figure:attention_type_2_selection}
    \end{figure}

\begin{figure}[h]
        \centering
        \includegraphics[width=0.45\textwidth]{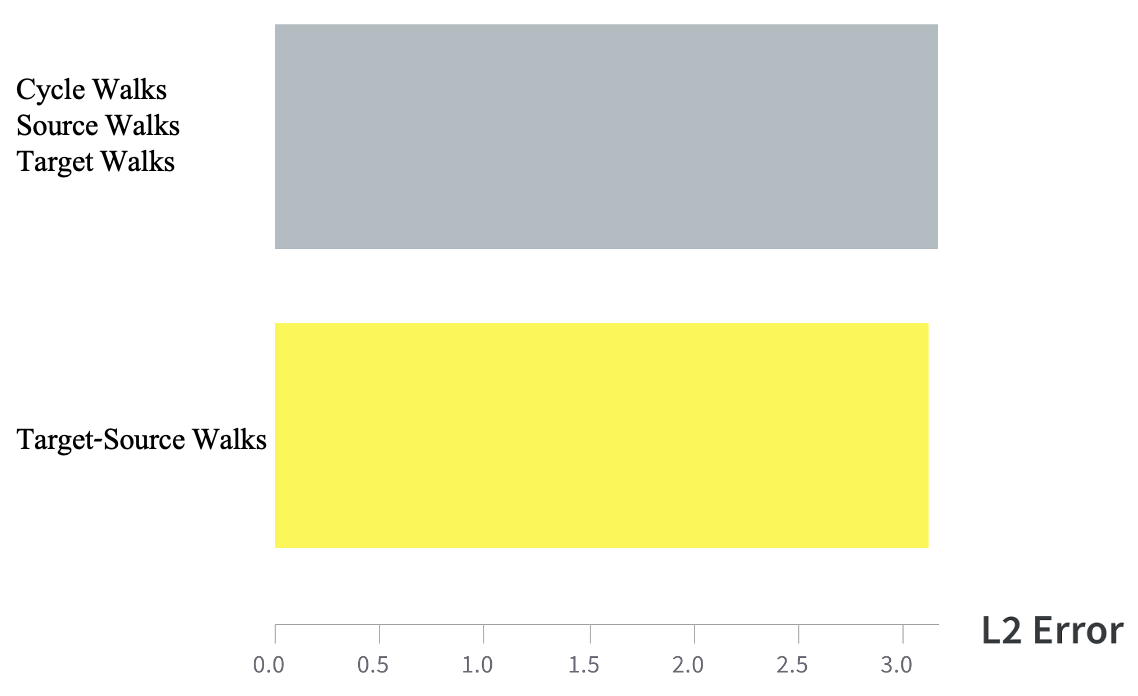}
        \includegraphics[width=0.45\textwidth]{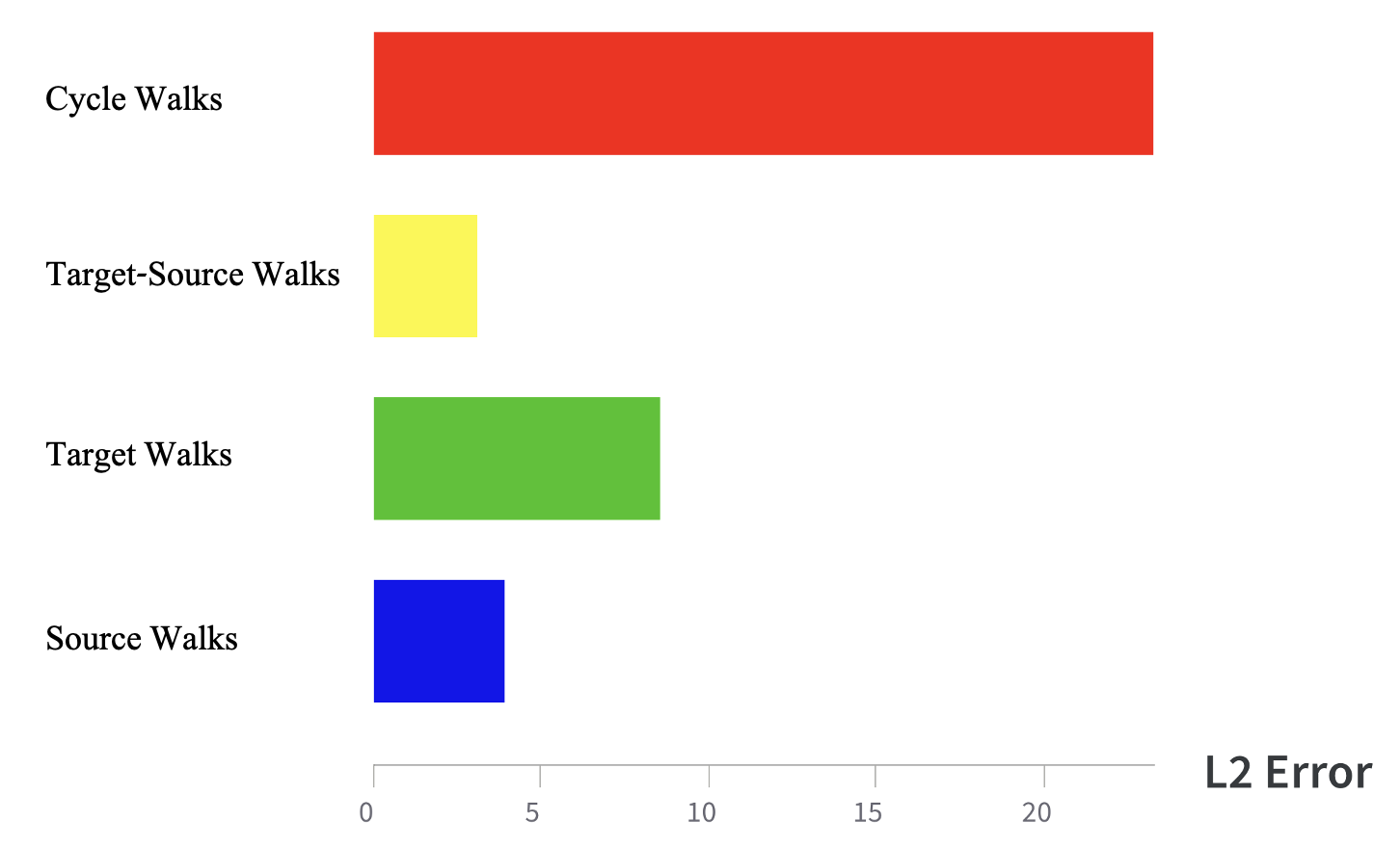}

        \caption{The L2 error of \acronym on the "Count walks starting and ending in the same red vertex" task,   tested with different walk types.} 

        \label{Figure:attention_type_3_selection}
    \end{figure}

\begin{figure}[h]
        \centering
        \includegraphics[width=0.45\textwidth]{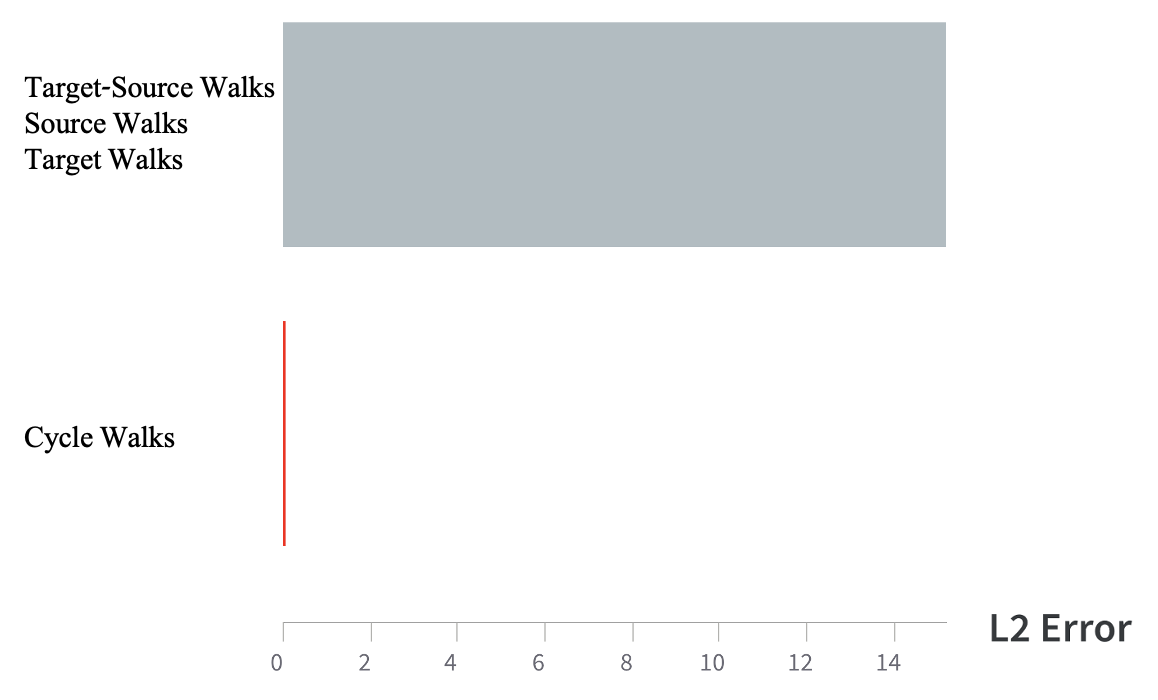}
        \includegraphics[width=0.45\textwidth]{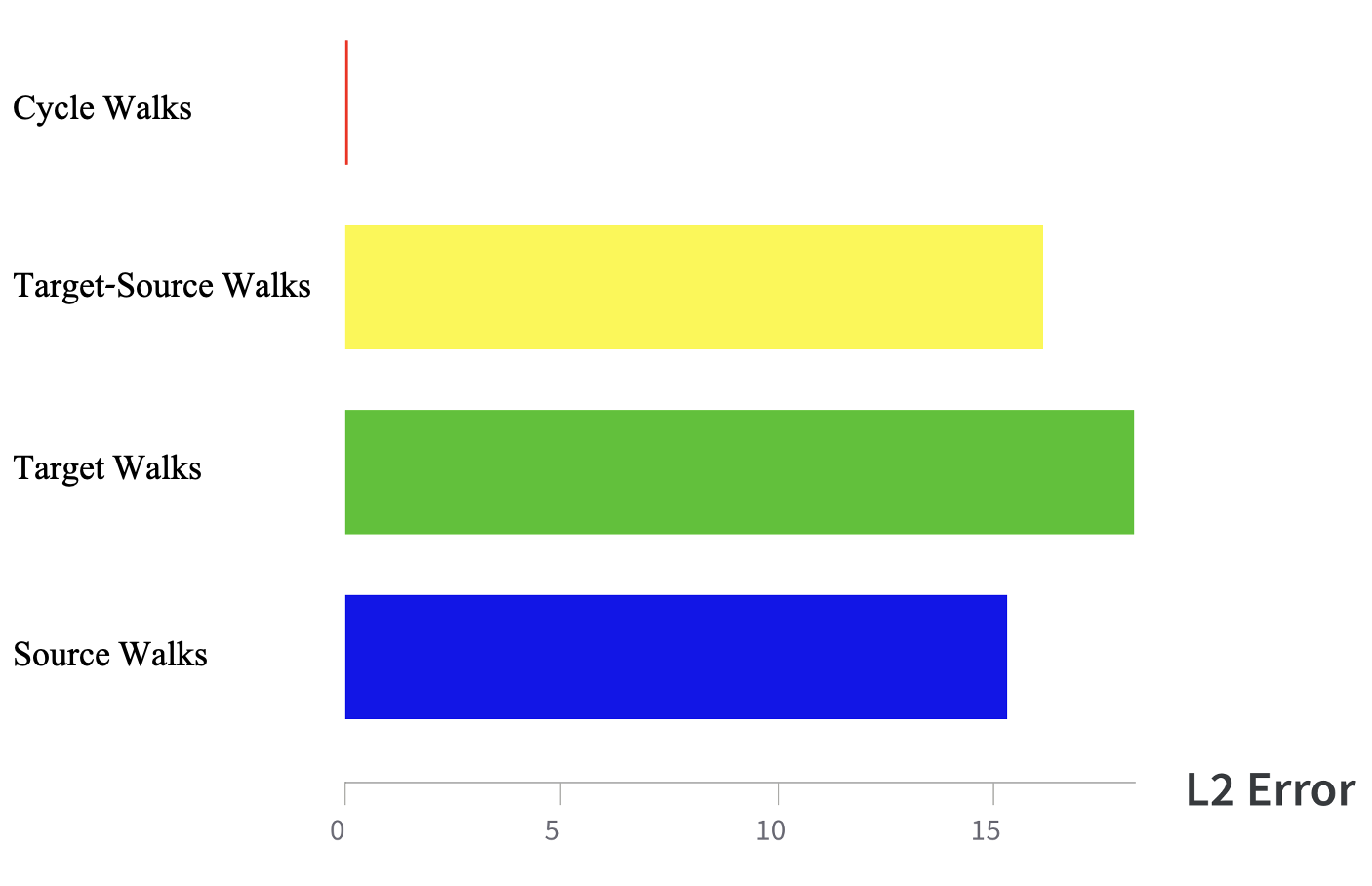}

        \caption{The L2 error of \acronym on the "Count walks between red vertices" task, tested with different walk types.} 

        \label{Figure:attention_type_4_selection}
    \end{figure}

As discussed in the main paper, there are many ways to limit walks to vertices. In \acronym we use $4$ ways that are natural and effective.
We now present $4$ synthetic tasks, showing that each of these walk types is superior for some task. On each task, we ran $5$ experiments. In $4$ experiment we used \acronym where we only allow one type of walks. Then, we compared the lowest loss to \acronym  when we allow $3$ types of walks: the $3$ types that did not have the lowest loss. The tasks are:
\begin{enumerate}
    \item Count walks starting in red vertices.
    \item Count walks starting and ending in the same red vertex.
    \item Count walks ending in red vertices.
    \item Count walks between red vertices.
    
\end{enumerate}
In Figures~\ref{Figure:attention_type_1_selection}, \ref{Figure:attention_type_2_selection},  \ref{Figure:attention_type_3_selection}, 
\ref{Figure:attention_type_4_selection} figures, each couple of figures corresponds to one task of the above tasks.  In each task, we plot the $4$ experiments where only one walk type is used (right subfigures) and the experiment where we allow all $3$ types of walk that did not achieve lowest error (left subfigures). 

%% file: invariance_lemma_proof.tex
\begin{lemma*}
\acronym is invariant to permutations in the case of graph labeling and equivariant in the case of vertex labeling.
\end{lemma*}

We first show that every component in the split criterion of \acronym is permutation-equivariant, except for the aggregation function applied in the case of graph-level tasks, which is permutation-invariant.
Let $\pi$ be a permutation over the vertices, and $P_\pi$ the corresponding permutation matrix.
For any permutation matrix $P$, it holds that $P^TP = I$, therefore:  
\begin{equation*}
\begin{aligned}
& \left(P_\pi AP_\pi^T \right)^d \left(P_\pi \f_k \right) 
 \\
&=\left(P_\pi AP_\pi^T \right)\left(P_\pi AP_\pi^T \right)
...\left(P_\pi AP_\pi^T \right)  \left(P_\pi \f_k \right)\\
&= P_\pi\left(A^d  \f_k\right)
\end{aligned}
\end{equation*}
And therefore $A^d  \f_k$ is permutation-equivariant.

Similarly, when subsets are used, if one uses a mask $M$, the above equivariance similarly holds because:
\begin{equation*}
\begin{aligned}
  &\left(\left(P_\pi A P_\pi^T \right)^d \odot \left(P_\pi M P_\pi^T \right)\right)  \left(P_\pi \f_k \right) \\
  & =\left(\left(\left(P_\pi AP_\pi^T \right)\left(P_\pi AP_\pi^T \right)...\left(P_\pi AP_\pi^T \right) \right)\odot \left(P_\pi M P_\pi^T \right)\right)  \left(P_\pi \f_k \right)\\
&=\left(\left(P_\pi A^d P_\pi^T \right) \odot \left(P_\pi M P_\pi^T \right)\right) \left(P_\pi \f_k \right)\\
& =\left(P_\pi \left(A^d \odot M\right) P_\pi^T \right)\left(P_\pi d \f_k \right) = P_\pi\left(\left(A^d \odot M\right) \f_k\right)  
\end{aligned}
\end{equation*}


Next, recall that in the case of vertex labeling, the threshold is applied to the vector $A^d  \f_k$ entry-wise. For every $\theta \in \reals$ it holds that $$ \left(A^d  \f_k\right)_i \geq \theta \iff P_\pi\left(A^d  \f_k\right)_{\pi(i)} \geq \theta $$
therefore thresholding  $A^d  \f_k$ is permutation-equivariant. Overall, in the case of vertex labeling, \acronym's split criterion is a decomposition of permutation-equivariant functions, and therefore it is permutation-equivariant.

In the case of graph labeling, the elements of the vector $A^d  \f_k$ are aggregated using a permutation-invariant function, and the threshold is applied to the aggregation result. As the vector $A^d  \f_k$ is permutation-equivariant, aggregating over its elements is permutation-invariant:

$$  AGG\left(\left(P_\pi AP_\pi^T \right)^d  \left(P_\pi \f_k \right)\right) = AGG\left(P_\pi\left(A^d  \f_k\right)\right)= AGG\left(A^d  \f_k\right)$$

Thresholding in this case is permutation-invariant, and therefore for graph labeling tasks \acronym's split-criterion is permutation-invariant.

%% file: gta_running_time_lemma_proof.tex
\begin{lemma*}
 The running time of searching the optimal split in \acronym is linear in the input parameters: maximal walk length , maximal number of ancestors to consider when searching for graph subsets, and number of features.
\end{lemma*} 

Below we compute how many dynamic features are considered at every split-node. Each dynamic feature is defined by the depth of the propagation $d$, an input feature $k$, and a subset of the vertices.
In graph labeling tasks the aggregation function chosen from one of $4$ options, is another parameter to consider. Therefore, the computational complexity depends on the number of possible combinations to these parameters.
 The possible values for $d$ are bounded by a hyper-parameter we denote here as $\textit{max\_depth}$. The number of subsets that are considered in a split-node $v$ is at most $2\textit{depth}(v) + 1 $ where $\textit{depth}(v)$ is the depth of the node $v$ in the tree. This can be further reduced by 
 a hyper-parameter $a$ that limits a split-node to only consider ancestors that are t distance st most $a$ from it. 
 This will reduce the number of subsets used to $2a+1$ at most. The walks are restricted to the subset in one of at most $4$ possible ways. Finally, we have just $4$ aggregation functions for graph-level tasks, and no aggregation is used in the case of vertex-level tasks.
Overall, in every split, \acronym computes at most $4\times 4 \times \left(2a+1\right)\times\textit{max\_depth}\times l$ dynamic features.
Therefore, the number of dynamic features in each split is linear in the number of input-features $l$, as in standard decision trees that only acts on the input features.
Notice that during inference it is possible to do a lazy evaluation of dynamic features such that only features that are actually used in the decision process are evaluated.

%% file: gta_geq_dt_lemma_proof.tex
     \begin{figure}[t]
    \centering
    \includegraphics[width=0.4\linewidth]{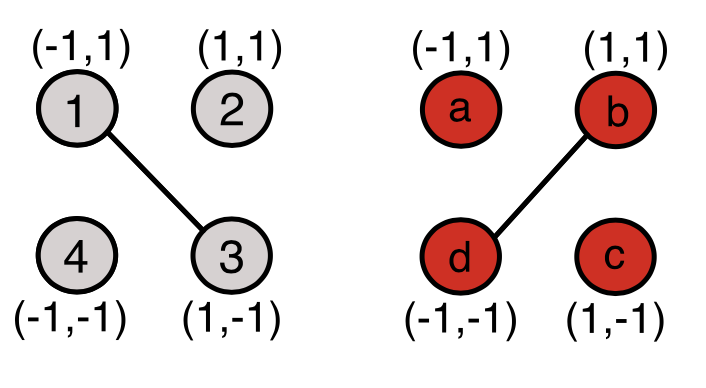}
    \caption{$G_1$(red) and $G_2$(grey) are two graphs on 4 vertices that are positioned on the plane at $\{\pm 1\} \times \{\pm 1\}$. Each graph has two features corresponding to its coordinates. In the proof of Lemma 4.3 we show that these graphs are indistinguishable by \acronym without subsets, but are separable when subsets are used.} 
    \label{Figure:lemma_graphs} 
\end{figure}

\begin{lemma*}
There exist graphs that are separable by \acronym , but are inseparable if \acronym is limited to not use subsets.
 \end{lemma*}

    

     Consider graphs on $4$ vertices that are positioned on the plane at $\{\pm 1\} \times \{\pm 1\}$ such that each vertex has two features $\f_1$ and $\f_2$ corresponding to its $x$-axis and $y$-axis. We consider two such graphs as presented in \Figref{Figure:lemma_graphs}.
     Note that for any of the features $A^0 f$ is identical for $G_1$ and $G_2$ since it ignores the edges of these graphs. For $d\geq1$ note that $A^d \f$ is a permutation of the vector $(1,-1,0,0)$ for $G_1$ and $G_2$. Since the aggregation functions are permutation invariant, applying any aggregation function to $A^d \f$ will generate the same value for $G_1$ and $G_2$. Moreover, since the topology of $G_1$ and $G_2$ is isomorphic, any invariant graph-theoretic feature would have the same value of $G_1$ and $G_2$. Therefore, \acronym without subsets cannot distinguish between $G_1$ and $G_2$. Nevertheless, \acronym can distinguish between them when using subsets.  Specifically, the tree shown in \Figref{Figure:dt_lemma} will separate these two graphs to two different leaves. The root of the tree conditions on $f_1$ which generates a subset of the vertices $\{(1,1), (1,-1)\}$. At the following level, the tree propagates $f_2$ along walks of length $2$ masking on this subset. For $G_1$, the only walk that is not masked out is the walk starting and ending in $(1,1)$ and therefore for this vertex the value $1$ is computed while $0$ is computed for all other vertices. However, for $G_2$, the only walk of length $2$ that is  not masked out is the one starting and ending at $(1,-1)$. Therefore, in $G_2$, the value $-1$ is computed for the vertex and $(1,-1)$ and the value $0$ is computed for all other vertices. See \Figref{Figure:dt_prop} for a visualization of this analysis.

Therefore, $(A^2\odot M_3(\{(1,1), (1,-1)\})\f_2$ is a permutation of the vector $(1,0,0,0)$ for $G_1$ and a permutation of the vector $(-1,0,0,0)$ for $G_2$ and thus \acronym can distinguish between these graphs.

%% file: gta_neq_gnns_lemma_proof.tex
 \begin{lemma*}
 There exist graphs that cannot be separated by GNNs but can be separated by \acronym.
\end{lemma*}

     \begin{figure}[t]
    \centering
    \includegraphics[width=0.5\linewidth]{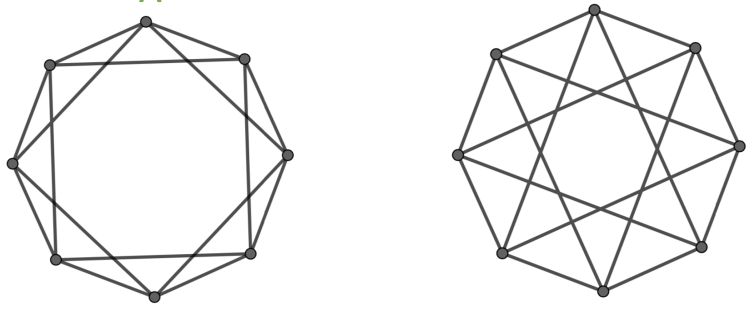}
    \caption{Two $4$-regular non-isomporphic graphs. The two graphs differ, for example, in their number of cycles of length $3$. The figure is taken from~\cite{idgnn}}.
    \label{Figure:regular_graph} 
\end{figure}


  Let $G_1$, and $G_2$ be the two $4$-regular non-isomorphic graphs as in \Figref{Figure:regular_graph}. Assume that all vertices have the same fixed feature $1$. Namely, we wish to distinguish these two graphs only by their topological properties. It is easy to see that a GNN will not be able to distinguish these two graphs, as after the message passing phase, all nodes will result in the same embedding~\cite{idgnn}. In contrast, \acronym is able to separate these two graphs with a single tree consisting of a single split node, by applying the split rule: $\sum(A^3 \odot M_t(V))f_1 \lessgtr 1$. This rule counts the number of cycles of length $3$ in each graph.
    Notice that GNNs here refer to the class of MPGNNs~\cite{mpgnn}, whose discriminative power is known to be bounded by the $1$-WL test~\cite{gin}. In particular, popular models such as GIN, GAT and  GCN~\cite{gin, gat, gcn}, which are also presented in the empirical evaluation sections, all belong to this class of GNNs. GIN~\cite{gin} with a sufficient amount of layers was shown to have exactly the same expressive power as 1-WL.

     \begin{figure}[ht]
    \centering
    \includegraphics[width=0.4\linewidth]{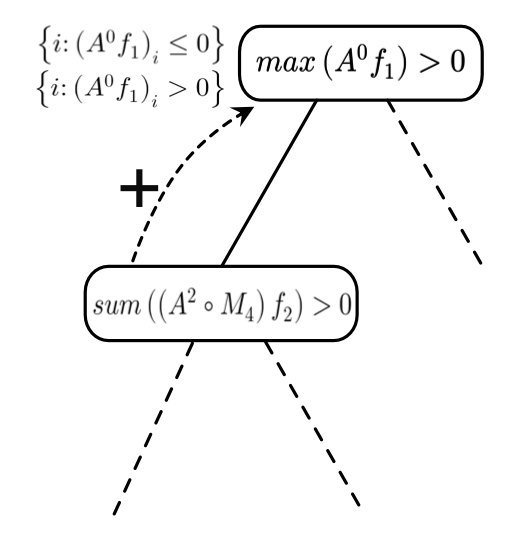}
    \caption{The TREE-G that separates the two graphs $G_1$ and $G_2$ in Figure \ref{Figure:lemma_graphs}.}
    \label{Figure:dt_lemma} 
\end{figure}

     \begin{figure*}[ht]
    \centering
    \includegraphics[width=0.7\linewidth]{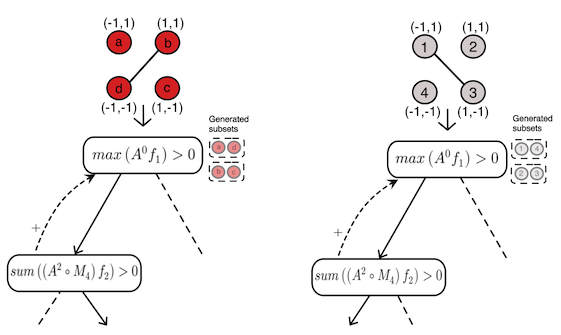}
    \caption{The prediction path of the graphs $G_1$(red) and $G_2$(grey) in the tree. The path is marked with bold arrows. The available subsets that are generated by the root are marked with a dashed line next to the root edges. The selected subset used in the inner split node is shown left of the node.}
    \label{Figure:dt_prop} 
\end{figure*}

%% file: tables/dataset_sum.tex
\begin{table*}[t]
  \centering

\begin{tabular}{l|ccccc} 
\\
Dataset  & \# Graphs & Avg \# Vertices & Avg \# Edges & \# Node Features &\# Classes  \\ 
   \toprule

Proteins~\cite{tudataset}   & 1113     & 39.06       & 72.82    &3   & 2          \\

Mutag~\cite{tudataset}    & 188      & 17.93       & 19.79      &7 & 2          \\

D\&D~\cite{tudataset}        & 1178     & 284,32      & 715.66  &89   & 2          \\

NCI1~\cite{tudataset}       & 4110     & 29.87       & 32.3     &37   & 2          \\ 
PTC~\cite{tudataset}        & 344      & 14          & 14      &19    & 2          \\

IMDb-B~\cite{tudataset}    & 1000     & 19          & 96     &0     & 2          \\ 
IMDb-M~\cite{tudataset}    & 1500     & 13          & 65    &0      & 3          \\ 

Mutagenicity~\cite{tudataset}        & 4337      & 30.32          & 30.37      &7    & 2          \\
Enzymes~\cite{tudataset}        & 600      & 32.63          & 62.14      &3    & 6          \\
molHIV~\cite{ogb}        & 41,127      & 25.5          & 27.5        &9  & 2\\ 
\bottomrule
\end{tabular}

\begin{tabular}{l|cccc} 
\\
Dataset & \# Vertices & \# Edges &\# Features &\# Classes  \\ 
    \toprule
Cora~\cite{planetoid}   & 2,708      & 10,556       & 1,433       & 7          \\ 

Citeseer~\cite{planetoid} & 3,327     & 9,104      & 3,703       & 6          \\ 

Pubmed~\cite{planetoid}       & 19,717     & 88,648      & 500      & 3          \\ 

Arxiv~\cite{ogb}   & 169,343     & 1,166,243          & 128          & 40          \\ 

Cornell~\cite{10.5555/295240.295725} & 183 & 295& 1,703&  5\\

Actor~\cite{Pei2020GeomGCN}& 7600& 33544& 931 & 5\\

Country~\cite{ivanov2021boost}& 3217 & 12684&  7& NA\\

    \bottomrule
\end{tabular}

  \caption{Statistics of datasets used in experiments.}
  \label{Table:data_sum}
\end{table*}